\def\year{2022}\relax
\documentclass[letterpaper]{article} 
\usepackage{aaai22}  
\usepackage{times}  
\usepackage{helvet}  
\usepackage{courier}  
\usepackage[hyphens]{url}  
\usepackage{graphicx}  
\usepackage{subfigure}
\usepackage{natbib}  
\usepackage{caption} 
\DeclareCaptionStyle{ruled}{labelfont=normalfont,labelsep=colon,strut=off} 

\usepackage[utf8]{inputenc}

\usepackage{amsmath}
\usepackage{booktabs}
\usepackage{algorithm}
\usepackage{algorithmic}
\usepackage{multirow}
\usepackage[pdfstartview=FitH,bookmarksnumbered=true,colorlinks,bookmarksopen=true]{hyperref}
\usepackage{newfloat}
\usepackage{listings}
\usepackage{xcolor}
\definecolor{mypink1}{rgb}{0.858, 0.188, 0.478}

\floatstyle{ruled}
\newfloat{listing}{tb}{lst}{}
\floatname{listing}{Listing}

\usepackage{mathtools}
\usepackage{amssymb}
\usepackage{amsthm}

\usepackage{scalerel,stackengine}
\stackMath
\newcommand\reallywidehat[1]{%
	\savestack{\tmpbox}{\stretchto{%
			\scaleto{%
				\scalerel*[\widthof{\ensuremath{#1}}]{\kern-.6pt\bigwedge\kern-.6pt}%
				{\rule[-\textheight/2]{1ex}{\textheight}}
			}{\textheight}%
		}{0.5ex}}%
	\stackon[1pt]{#1}{\tmpbox}%
}

\title{Sim2Real Object-Centric Keypoint Detection and Description}

\author{Paper ID [4781]}

\author{
	Chengliang Zhong\thanks{These authors contributed equally. Corresponding author: Fuchun Sun.}$^{12}$,
	Chao Yang\footnotemark[1]$^2$,
	Jinshan Qi$^4$,
	Fuchun Sun$^2$,\\
	Huaping Liu$^2$,
	Xiaodong Mu$^1$,
	Wenbing Huang$^3$
	}
\affiliations{
    \textsuperscript{\rm 1}Xi'an Research Institute of High-Tech, Xi'an 710025, China\\ 
	\textsuperscript{\rm 2}Beijing National Research Center for Information Science and Technology (BNRist),\\ State Key Lab on Intelligent Technology and Systems,\\ Department of Computer Science and Technology, Tsinghua University, Beijing 100084, China\\
	\textsuperscript{\rm 3} Institute for AI Industry Research, Tsinghua University \\
	\textsuperscript{\rm 4}Shandong University of Science and Technology, Qingdao 266590, China\\
	\texttt{zhongcl19@mails.tsinghua.edu.cn, fcsun@tsinghua.edu.cn}\\
}

\begin{document}
\maketitle
\begin{abstract}

Keypoint detection and description play a central role in computer vision. Most existing methods are in the form of \emph{scene-level} prediction, without returning the object classes of different keypoints. In this paper, we propose the \emph{object-centric} formulation, which, beyond the conventional setting, requires further identifying which object each interest point belongs to. 
With such fine-grained information, our framework enables more downstream potentials, such as object-level matching and pose estimation in a clustered environment. To get around the difficulty of label collection in the real world, we develop a sim2real contrastive learning mechanism that can generalize the model trained in simulation to real-world applications. 
The novelties of our training method are three-fold: (i) we integrate the uncertainty into the learning framework to improve feature description of hard cases, e.g., less-textured or symmetric patches; (ii) we decouple the object descriptor into two output branches---intra-object salience and inter-object distinctness, resulting in a better pixel-wise description; (iii) we enforce cross-view semantic consistency for enhanced robustness in representation learning. Comprehensive experiments on image matching and 6D pose estimation verify the encouraging generalization ability of our method from simulation to reality. Particularly for 6D pose estimation, our method significantly outperforms typical unsupervised/sim2real methods, achieving a closer gap with the fully supervised counterpart. Additional results and videos can be found at \href{https://zhongcl-thu.github.io/rock/}{https://zhongcl-thu.github.io/rock/}.

\end{abstract}
	
	\section{Introduction}\label{sec:intro}
	
	Extracting and describing points of interest (\emph{keypoints}) from images are fundamental problems for many geometric computer vision tasks such as image matching~\cite{sift}, camera calibration~\cite{calibration}, and visual localization~\cite{PIASCO201890}. Particularly for image matching, it requires searching the same and usually sparse keypoints for a pair of images that record the same scene but under a different viewpoint.

	\begin{figure}[H]
        \centering
        \includegraphics[height=0.31\textwidth]{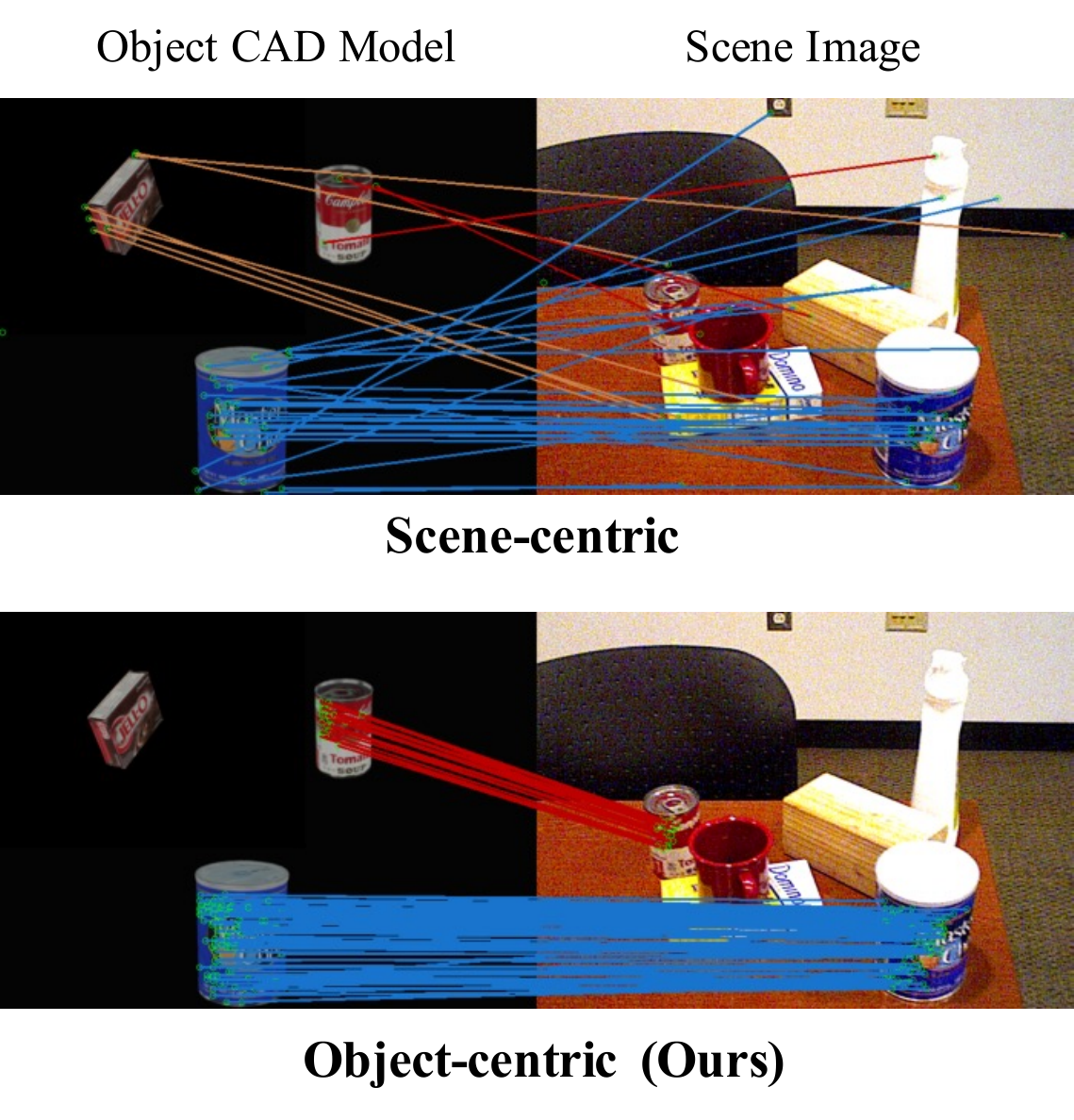}
        \caption{
        Keypoint matching processing for scene-centric method, R2D2~\cite{r2d2} and our object-centric method. Our method accurately matches the keypoints on different objects, while R2D2 predicts some unwanted points located in the background.}
        \label{fig:overall}
    \end{figure}
    
	A variety of works have been done towards keypoint detection and description, ranging from traditional hand-crafted methods~\cite{sift,surf} to current data-driven approaches~\cite{superpoint,r2d2,disk}. Despite the fruitful progress, most existing methods are initially targeted on image-level/scene-centric tasks, making them less sophisticated for other more fine-grained problems, \emph{e.g.}, the object-level matching or pose estimation. Object-level tasks are crucial in many applications. A typical example is in robotic grasp manipulation, where for planning a better grasp posture, the robot requires to compare the keypoints between the scene image and the CAD model followed by estimating the pose of the object according to the keypoint correspondence~\cite{sparse}.
	If we apply the previous methods (such as R2D2~\cite{r2d2}) straightly, the detected keypoints from the scene image usually contain not only the desired points on the target object, but also those unwanted points located in the background that share a similar local texture with the object CAD model, as illustrated in Figure.~\ref{fig:overall} (top row). Such failure is natural as R2D2 never tells which keypoints are on the same object, and it solely captures the local similarity therein.

	In this paper, we propose a novel conception: \emph{object-centric} keypoint detection and description, in contrast to the conventional \emph{scene-centric} setting. Beyond keypoint detection and description, the proposed object-centric formulation further teaches the algorithm to identify which object each keypoint belongs to. Via this extra supervision, our formulation emphasizes the similarity between keypoints in terms of both local receptive field, and more importantly, objectness. Figure.~\ref{fig:overall} depicts that the object-centric method (bottom row) accurately predicts the object correspondence (different colors) and matches the keypoints on different objects between the scene image and the CAD model, thanks to the object-wise discrimination.
	 
	Obtaining object annotations, of course, is resource-consuming in practice. Yet, we find it well addressable if leveraging the  sim2real training and domain randomization  mechanism~\cite{se3}. By this strategy, we can easily obtain rich supervision and manipulate the training samples arbitrarily to serve our goal. Specifically in our scenario, we can access the transformation projection between different views of the same scene image as well as the object label of each pixel. Based on the richly annotated simulation data, we develop a contrastive learning framework to jointly learn keypoint detection and description. To be specific, it contains three main components: (i) A novel uncertainty term is integrated into object keypoints detection, which is expected to handle challenging objects with similar local textures or geometries. (ii) As for the keypoint description, it is divided into two output branches, one to obtain intra-object salience for the keypoints on the same object and the other to enforce inter-object distinctness across the keypoints on different objects. (iii) For each target object in the scene image, a contrastive sample containing only the object of the same view but with a clean background is also adopted to derive better cross-view semantic consistency. Once the model is trained in simulation, it can be applied to real applications.
	
	We summarize our contributions below.
	\begin{itemize}
	    \item To the best of our knowledge, we are the first to raise the notion of object-centric keypoint detection and description, which better suits the object-level tasks.
	    \item To address the proposed task, we develop a novel sim2real training method, which enforces uncertainty, intra-object salience/inter-object distinctness, and semantic consistency.
	    \item Comprehensive experiments on image matching and  6D  pose estimation verify the encouraging generalization ability of our method from simulation to reality.
	\end{itemize}

	\section{Related Work}\label{sec:related_work}
	Here, we focus on local keypoint detection and description from a 2D image. 
    \paragraph{\textbf{Scene-centric method.}} 
    The hand-crafted methods often employ corners~\cite{susan,fast} or blobs~\cite{surf} as keypoints whose associated descriptions are based on histograms of local gradients, including the famous SIFT descriptor~\cite{sift}. 
    In part, this is due to the increased complexity of scene semantics, which exacerbates the reliance of keypoint detection and description to modern deep learning approaches. Such data-driven methods can be categorized into learned detectors~\cite{tilde, key.net}, learned descriptors~\cite{matchnet,balntas2016pnnet}, and the combination of them~\cite{yi2016lift,d2net,r2d2}.
    
    
    \paragraph{\textbf{Object-centric keypoint detection.}}
      According to the different levels of supervision,
      object-centric keypoint detection can be divided into fully supervised~\cite{bb8, pvnet}, semi-supervised~\cite{s3k}, and self-supervised learning methods~\cite{okpose, 2019nips_obj_kp}. To distinguish the objectness of keypoints, most researchers utilize pre-trained object detectors to focus on small patches of different objects and find keypoints for each. Also, the number of keypoints of each object would be fixed. Although some works~\cite{s3k} did not involve the object detector, there is only one object in each image. These methods are difficult to generalize to new objects because of the specialized object detectors and fixed number of keypoints. 
      In the keypoints detection part, the unsupervised learning method~\cite{2019nips_obj_kp} will predict the keypoints of multiple objects at once. However, their methods are limited to simple scenarios. 
    
    \paragraph{\textbf{Object-centric keypoints descriptors.}} 
    Dense-Object-Net(DON)~\cite{don} is the first work to learn the dense descriptors of objects in a self-supervised manner. Based on DON, MCDONs~\cite{chai2019multi} introduces several contrastive losses to maintain inter-class separation and intra-class variation. However, a non-disentangled descriptor with two kinds of distinctiveness requires more supervision and increases the difficulty of model convergence. BIND~\cite{bind} uses multi-layered binary nets to encode edge-based descriptions, which is quite different from the point-based methods. Although the work of semantic correspondence~\cite{Yang2017ObjectAwareDS, sfnet} learns object-centric semantic representation, they directly predict dense correspondence from different views instead of outputting the high-level descriptors. To the best of our knowledge, jointly learned object-centric keypoint detection and description has not been explored before.

	\section{Object-centric Detection and Description}\label{sec:method}

    This section presents the problem formulation of the object-centric keypoint detection and description. The details of the sim2real contrastive training method are provided as well.

	\subsection{Formulation and over-all architecture}\label{sec:architecture}
	
	Keypoint detection and description is actually an image-based dense prediction problem. It needs to detect whether each pixel (or local patch) in the input image corresponds to the interest point or not. Besides detection, the predicted description vector of each pixel is essential, upon which we can, for example, compute the similarity between different keypoints in different images. Moreover, this paper studies the object-centric formulation; hence we additionally associate each descriptor with the objectness---the keypoints on the same object are clustered while those on different objects are detached from each other. The formal definitions are as below.
	
	\noindent
	\textbf{Keypoint detector.} 
	Given an input image $I\in\mathbb{R}^{3\times H\times W}$, the detector outputs a non-negative confidence map $\sigma(I)\in\mathbb{R}_{\geq 0}^{H\times W}$, where $H$ and $W$ are respectively the height and width. The pixel  of the $i$-th row and $j$-th column denoted as $I[i,j]$ will be considered as a keypoint if $\sigma(I)[i,j]>r_{\text{thr}}$ for a certain threshold $r_{\text{thr}}>0$. 
	
	\noindent
	\textbf{Keypoint descriptor with objectness.}
	The descriptor is formulated as $\eta(I)\in\mathbb{R}^{C\times H\times W}$, where $C$ represents the dimensionality of the descriptor vector for each pixel. As mentioned above, there are two subparts of each description vector, one for intra-object salience and the other one for inter-object distinctness. We denote them as $\eta_s(I)\in\mathbb{R}^{C_1\times H\times W}$ and $\eta_c(I)\in\mathbb{R}^{C_2\times H\times W}$, respectively, where $C_1+C_2=C$.

\begin{figure}[H]
        \centering
        \includegraphics[width=0.47\textwidth]{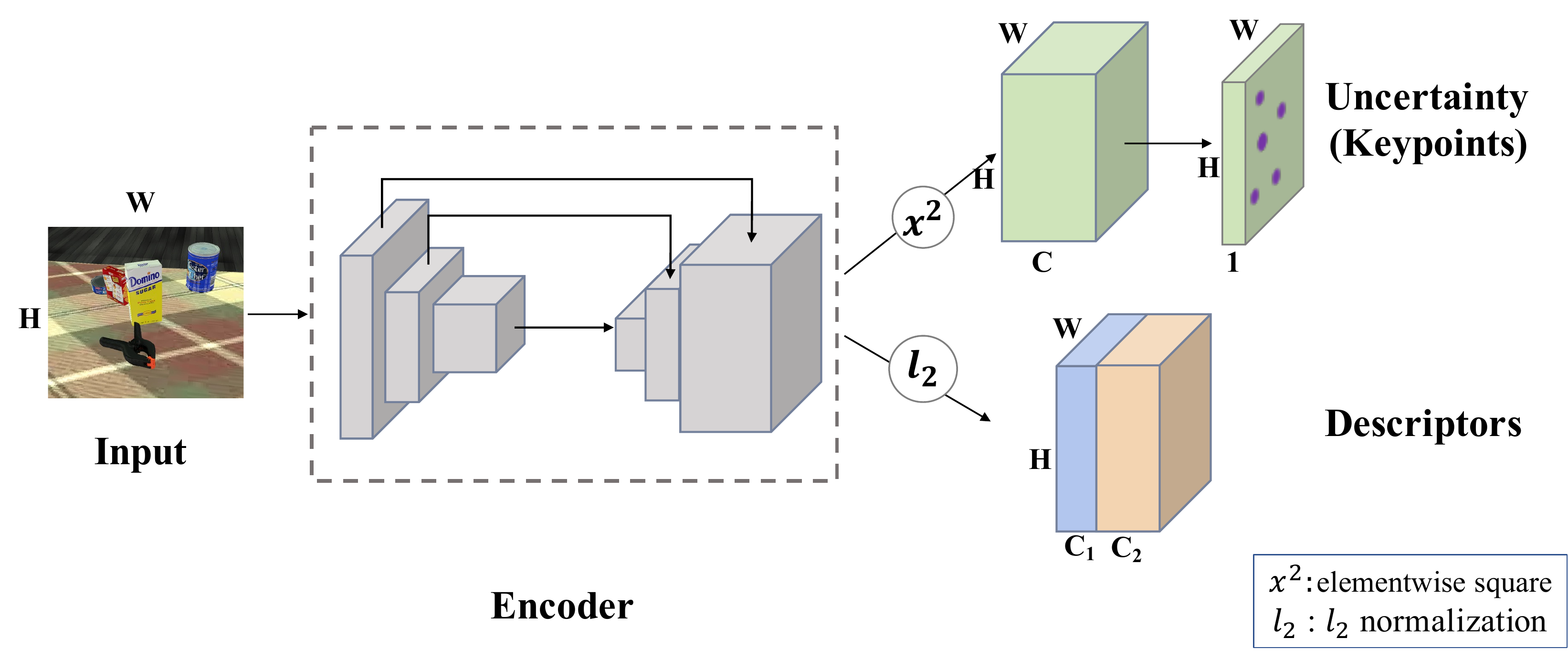}
        \caption{Overview of our network. The uncertainty map is associated with the detector and the descriptor is disentangled into two parts which respectively pursue inter-object salience and inter-object distinctness.}
        \label{fig:architecture}
    \end{figure}

	Similar to previous works (such as R2D2), the detector and descriptor share a major number of layers, called an encoder. For the implementation of the encoder, we apply Unet~\cite{unet} plus ResNet~\cite{resnet} blocks and upsampling layers as the backbone, inspired by Monodepth2~\cite{monodepth2} which is originally for depth estimation. We have also made some minor modifications regarding the encoder to deliver better expressivity. The output of the encoder serves as input to i) the detector after element-wise square operation and ii) the descriptor after $\ell_2$ normalization layer, motivated by design in R2D2~\cite{r2d2}. 
	More implementation details are provided in the appendix. The overall architecture is illustrated in Figure~\ref{fig:architecture}.
    
    The training of our model is conducted in a simulated environment. In general, for each query image $I_1$, we collect its viewpoint-varying version $I_2$ (which is indeed the adjacent frame of $I_1$ as our simulation data are videos). Besides, we generate a rendered version of $I_1$ by retaining the target object only and cleaning all other objects and background; we call this rendered image $I_0$. With these three images, we have the overall training loss as follows.
    \begin{align}\label{formula:L_total}
        \mathcal{L}(I_1, I_2, I_0)=\mathcal{L}_{r}(I_1, I_2)&+\lambda_1 \mathcal{L}_{d_s}(I_1, I_2, I_0) \nonumber \\ &+ \lambda_2 \mathcal{L}_{d_c}(I_1, I_2, I_0)\text{,}
    \end{align}
    where $\mathcal{L}_r$ stands for enforcing the repeatability of keypoint detector, $\mathcal\mathcal{L}_{d_s}$ and $\mathcal{L}_{d_c}$ are the objectives of the descriptor for the intra-object salience and inter-object distinctness, respectively, $\lambda_1$ and $\lambda_2$ are the associated trade-off weights.
    
    Now we introduce each loss. The motivation of minimizing $\mathcal{L}_r$ is to enforce the activation of the detector to be invariant with respect to the change of viewpoint, which is dubbed as repeatability by~\cite{r2d2}. Here, we introduce a metric operator $r$ different from that used in~\cite{r2d2}, which combines SSIM~\cite{ssim} and $\ell_1$ normalization:
    \begin{align}\label{formula:L_sim}
        r(x, y)=\frac{\alpha}{2}\left(1-\operatorname{SSIM}\left(x, y\right)\right)+(1-\alpha)\left\|x-y\right\|_{1}\text{ ,}
    \end{align}
    where $\alpha= 0.85$ by default. We thereby compute $\mathcal{L}_{r}(I_1, I_2)$:
    \begin{align}\label{formula:L_sim_us}
        \mathcal{L}_{r}(I_1, I_2) = \frac{1}{|\mathbb{U}_1|} \sum_{u_1\in \mathbb{U}_1}r\left(\sigma(I_1)[u_1], \sigma(I_2)[T_{12}(u_1)]\right),
    \end{align}
    where $\mathbb{U}_1$ refers to all $N\times N$ patches around each coordinate in $I_1$, $|\mathbb{U}_1|$ is the size of $\mathbb{U}_1$, $T_{12}$ is the coordinate transformation between $I_1$ and $I_2$, and thus $T_{12}(u_1)$ returns the corresponding coordinate of $u_1$ in $I_2$. 

    Both $\mathcal{L}_{d_s}$ and $\mathcal{L}_{d_c}$ are formulated by the contrastive learning strategy~\cite{infonce}. The only difference lies in the different construction of positive-negative training samples. By omitting the involvement of the rendered sample $I_0$, we first discuss the general form with uncertainty in~\ref{sec:jointly_learning} and then specify the difference between $\mathcal{L}_{d_s}$ and $\mathcal{L}_{d_c}$ in~\ref{sec:descriptor_disentangle}. In~\ref{sec:descriptorsemantic}, we further consider the training by adding $I_0$.

    \subsection{Contrastive learning with uncertainty}\label{sec:jointly_learning}

    We assume the general form of $\mathcal{L}_{d_s}(I_1,I_2)$ and $\mathcal{L}_{d_c}(I_1,I_2)$ (without $I_0$) to be $\mathcal{L}_c(I_1,I_2)$. A crucial property of keypoint descriptor is that it should be invariant to image transformations between $I_1$ and $I_2$ like viewpoint or illumination changes. We thus treat the descriptor learning as a contrastive learning task~\cite{wang2021dense}. To be specific, we define the query vector in $\eta(I_1)$ as $d_1$, the positive and negative vectors in $\eta(I_2)$ as $d_2^{+}$ and $d_2^{-}$, respectively. According to the definition in~\cite{infonce}, the contrastive loss with one positive sample $d_2^+$ and the negative set $\mathbb{D}_2^-$ is given by
    \begin{align}\label{formula:infonce}\begin{split}
    &\mathcal{L}_c(d_1, d_2^+, \mathbb{D}_2^-) =\\ &-\log\frac{\exp(d_1\cdot d_2^+/\tau)}{\exp(d_1\cdot d_2^+ /\tau)+\sum\limits_{d_2^-\in \mathbb{D}_2^-}{\exp(d_1\cdot d_2^-/\tau)}} \text{,}
    \end{split}
    \end{align}
    where $\tau$ is the temperature parameter.

    \begin{figure}[!ht]
        \centering
        \includegraphics[width=0.45\textwidth]{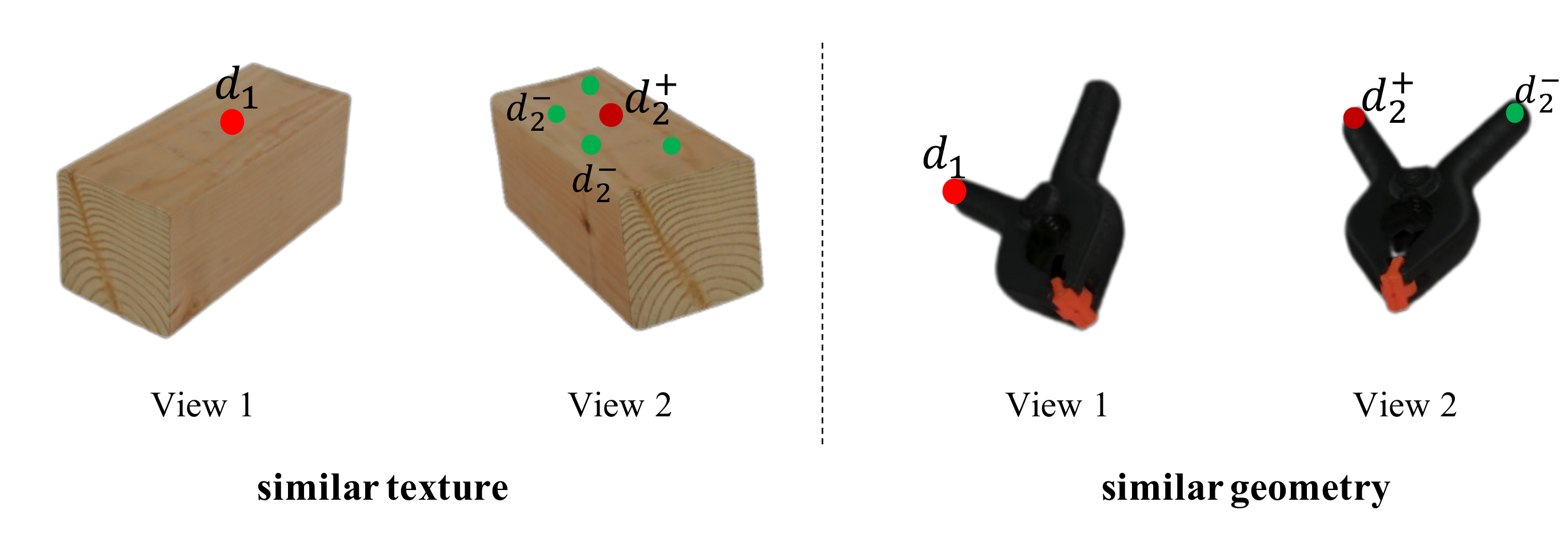}
        \caption{Less-textured and symmetric regions.}
        \label{fig:keypoint}
    \end{figure} 
    
    The quality of the contrastive samples influences the training performance greatly. Considering some objects with texture-less or symmetry geometries in Figure~\ref{fig:keypoint}, if the coordinate of $d^-_2$ is distant to $d_1$ after view projection, then $d^-_2$ is considered as a negative sample (the detailed negative sampling is in~\ref{sec:descriptor_disentangle}). Then, under the contrastive learning Eq.~\eqref{formula:infonce}, $d^-_2$ is enforced to be dissimilar to $d_1$ as much as possible. However, this conflicts with the texture/geometry distribution of the objects, as the local fields of $d_1$ and $d_2^-$ are indeed similar. We should design a certain mechanism to avoid such inconsistency. 
    Recall that our focus is on selecting sparse points of interest. The keypoint detector $\sigma(I_1)$ returns the confidence to determine which point should be selected. If we use this confidence to weight the importance of the query sample, we are more potential to filter the unexpected cases as mentioned above. 
    
    More specifically, we borrow the uncertainty estimation from~\cite{poggi, uncertain}, which has been proved to improve the robustness of deep learning in many applications. Starting by predicting a posterior $p(\mu|\bar{\mu}, \gamma)$ for the descriptor of each pixel parameterized with its mean $\mu$ and variance $\gamma$ over ground-truth labels $\mu$~\cite{d3vo}, the negative log-likelihood becomes:
    \begin{align}\label{formula:log}
        -\log p(\mu|\bar{\mu}, \gamma)=\frac{|\mu-\bar{\mu}|}{\gamma}+\log{\gamma} \text{.}
    \end{align}
    We adjust this formula into our contrastive learning framework. We first regard the reciprocal of the detection confidence as the uncertainty variance for each pixel, \emph{i.e.} $\gamma=\sigma(I_1)^{-1}$. It is reasonable as the larger confidence it outputs, the smaller uncertainty it exhibits. Second, we replace the error $|\mu-\bar{\mu}|$ with our loss $\mathcal{L}_c$ in Eq.~\eqref{formula:infonce}, since our goal is to refine the contrastive learning in the first place. 
    
    By summation over all queries in $I_1$, we derive:
    \begin{align}
    \label{formula:L_C}
        \mathcal{L}_d(I_1, &I_2) &=\frac{1}{M}\sum_{i=1}^M{\frac{\mathcal{L}_c(d^i_1, d^{i+}_2, \mathbb{D}^{i-}_2)}{(\sigma_1^i)^{-1}}+\log{(\sigma_1^i)^{-1}}} \text{,}
    \end{align}
    
    where, for the $i$-th query in $I_1$, $d^i_1$ indicates the description query vector, $d^{i+}_2$ and $\mathbb{D}^{i-}_2$ are the corresponding positive sample and negative sample set in $I_2$, ${\sigma}_1^i$ is the detection value, $M$ is the number of all queries.

    \subsection{Disentangled descriptor learning} \label{sec:descriptor_disentangle}
    As depicted in Figure~\ref{fig:architecture}, the descriptor is learned for two goals: it should not only distinguish different keypoints on the same object but also classify those across different objects. We realise this via two disentangled losses $\mathcal{L}_{d_s}(I_1,I_2)$ and $\mathcal{L}_{d_c}(I_1,I_2)$, which respectively follow the general form of $\mathcal{L}_d(I_1, I_2)$ in Eq.~\eqref{formula:L_C} and $\mathcal{L}_c$ in Eq.~\eqref{formula:infonce}. The two losses also employ distinct constructions of the training samples $d_2^+$ and $\mathbb{D}_2^-$ for any given query $d_1$. For better readability, we refer to their constructions as:  $d_{2,s}^+$ and $\mathbb{D}_{2,s}^-$, $d_{2,c}^+$ and $\mathbb{D}_{2,c}^-$, respectively. 
    
    \begin{figure}
        \centering
        \includegraphics[width=0.35\textheight]{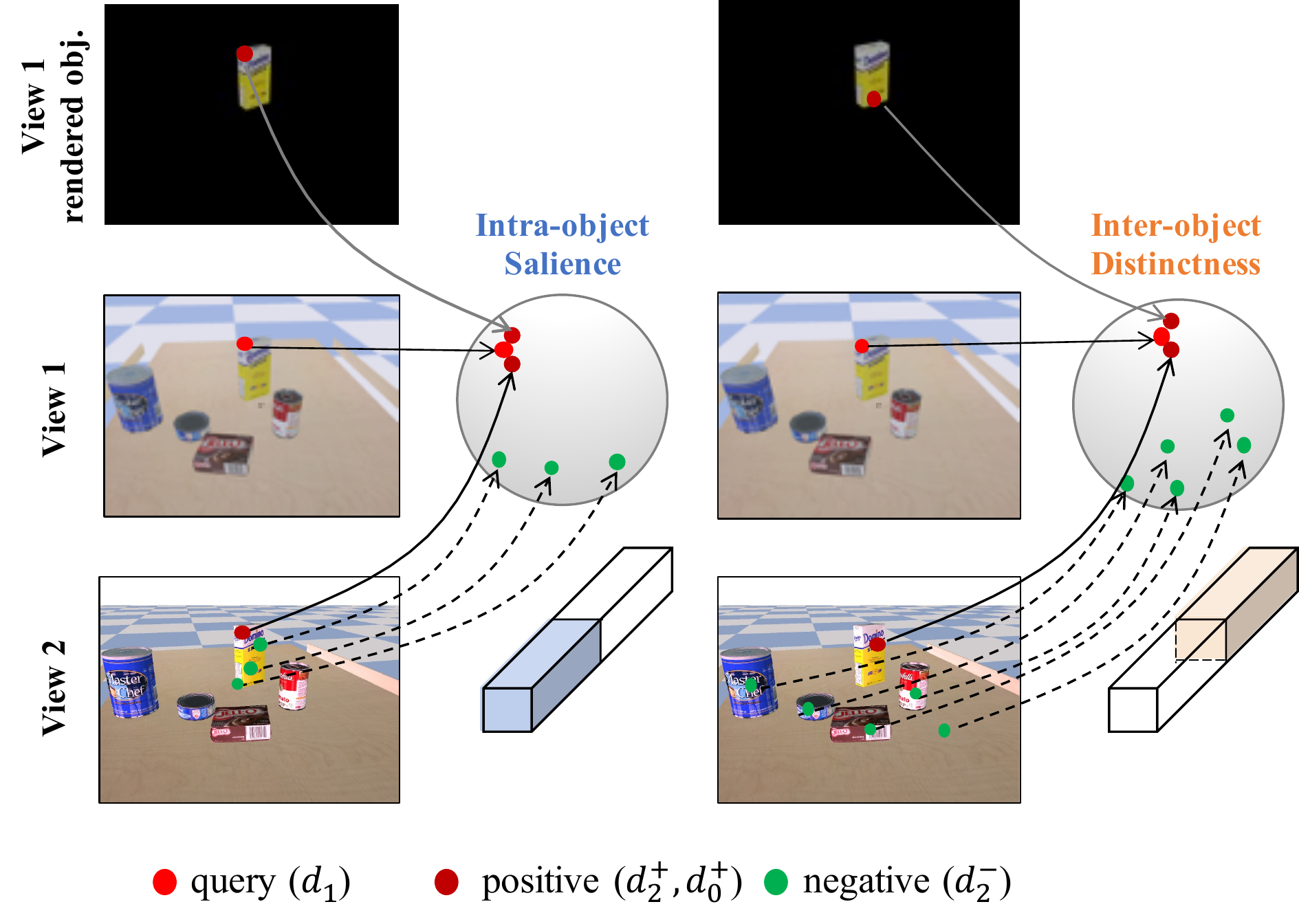}
        \caption{Illustration of the positive and negative samples for a given query. The middle and bottom rows denote the synthetic scenes from two different viewpoints. The top line renders the object with a clean background at view 1. We decouple the descriptor into two parts for learning the intra-object salience (first column) and the inter-object distinctness (second column).}
        \label{fig:pipeline}
    \end{figure}
    
    \noindent
    \textbf{Intra-object salience.}
    The loss $\mathcal{L}_{d_s}(I_1,I_2)$ is for intra-object salience. Suppose the coordinate of the query $d_1$ in image $I_1$ to be $u_1$, indicating $d_1=\eta_s(I_1)[u_1]$ where $\eta_s$ outputs the salience part of the descriptor as defined before. The positive sample $d_{2,s}^+$ is chosen as the projection from $I_1$ to $I_2$ via the view transformation $T_{12}$; in other words $d_{2,s}^+=\eta_s(I_2)[T_{12}(u_1)]$.  As for the negative candidates $\mathbb{D}_{2,s}^-$, we pick the points from $I_2$ on the same object as the query but out of the $\delta$-neighbourhood, namely, $\mathbb{D}_{2,s}^-=\{\eta_s(I_2)[u]\mid \ ||u-T_{12}(u_1)||_2>\delta, l(u)=l(T_{12}(u_1))\}$ where $l(u)$ returns the object label at pixel $u$. Figure~\ref{fig:pipeline} illustrates the sampling process (first column). By iterating over all possible queries, we arrive at the similar form to Eq.~\eqref{formula:L_C}:
    \begin{align}
        \label{formula:infonce_spatial}
        \mathcal{L}_{d_s}(I_1,I_2)&=\frac{1}{M}\sum_{i=1}^M{\frac{\mathcal{L}_c(d^i_1, d^{i +}_{2,s}, \mathbb{D}^{i -}_{2,s})}{(\sigma_1^i)^{-1}}+\log{(\sigma_1^i)^{-1}}} \text{,}
    \end{align}
    where $d^{i +}_{2,s}$ and $\mathbb{D}^{i +}_{2,s}$ are the positive sample and negative set for the $i$-th query. Via Eq.~\eqref{formula:infonce_spatial}, our hope is to accomplish the distinctness between keypoints on the same object. 
    
    \noindent
    \textbf{Inter-object distinctness.}
    We now introduce how to create the samples for $\mathcal{L}_{d_c}(I_1,I_2)$. Different from the above inter-object loss, here any point on the same object as the query in $I_2$ is considered as the positive sample, that is, $d_{2,c}^+\in \{\eta_c(I_2)[u]\mid \ l(u)=l(T_{12}(u_1))\}$ where $\eta_c$ denotes the inter-object output branch of the descriptor. In terms of the negative samples, the points on other objects or the background are selected, implying $\mathbb{D}_{2,c}^-=\{\eta_c(I_2)[u]\mid \ l(u)\neq l(T_{12}(u_1))\}$. The illustration is displayed in Figure~\ref{fig:pipeline} (second column). In form, we have the summation over all queries as follows:
    \begin{align}
        \label{formula:infonce_category}
        \mathcal{L}_{d_c}(I_1,I_2)&=\frac{1}{M}\sum_{i=1}^M{\mathcal{L}_c(d^i_1, d^{i +}_{2,c}, \mathbb{D}^{i -}_{2,c})} \text{.}
    \end{align}
    
    By making use of Eq.~\eqref{formula:infonce_spatial} and~\eqref{formula:infonce_category} together, we obtain more fine-grained information: we can tell if any two keypoints are on the same object, and if yes, we can further know if they correspond to different parts of the object by comparing their intra-descriptors.

    \subsection{Semantic consistency}\label{sec:descriptorsemantic}
\begin{figure}[!ht]
        \centering
        \includegraphics[width=0.4\textwidth]{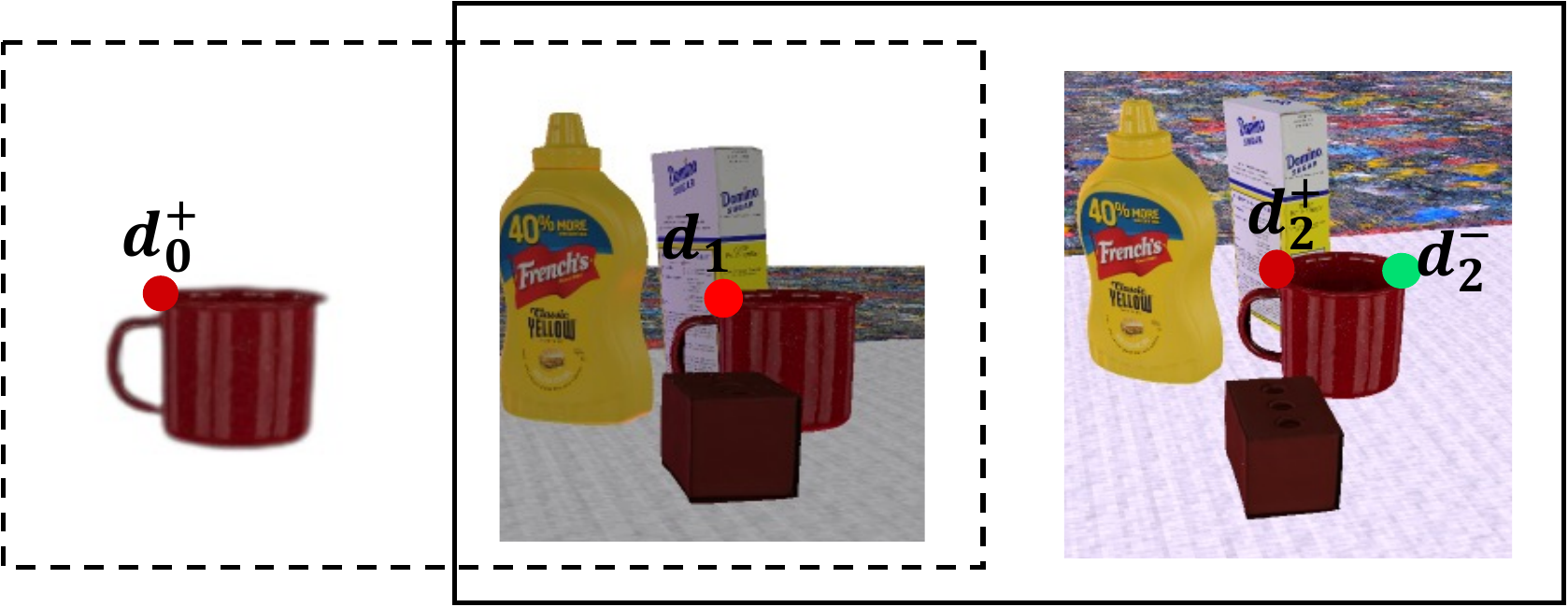}
        \caption{Illustration of semantic consistency paradigm.}
        \label{fig:semantic}
    \end{figure}

    This subsection presents how to involve the rendered image $I_0$ into our contrastive training. Selvaraju et al.~\cite{cast} found the existing contrastive learning models often cheat by exploiting low-level visual cues or spurious background correlations, hindering the expected ability in semantics understanding. This actually happens in our case when the descriptor may use the spatial relationship with other objects or the background. To illustrate this issue, in Figure~\ref{fig:semantic}, we assume the left (red) and right (green) edges of the cup's mouth are a pair of negative samples, their main difference lies in the distance from the handle of the cup. Nevertheless, the local region of the sugar box (yellow box) behind the red dot could be used as a shot-cut reference for the distinctness between the red and green points, which is NOT what we desire. The ideal learning of the descriptor for the object is to make it focus on the object itself. 
    
    For this purpose, we render the cup according to its pose in $I_1$ and remove all other things, leading to the image $I_0$. We then perform contrastive learning by further taking the positive from $I_0$ into account following the similar process in Eq.~\eqref{formula:infonce_spatial} and~\eqref{formula:infonce_category}. The whole pipeline is demonstrated in Figure~\ref{fig:pipeline}. The losses $\mathcal{L}_{d_s}(I_1,I_2)$ and $\mathcal{L}_{d_c}(I_1,I_2)$ are rewritten as follows.
    \begin{align}
        \label{formula:infonce_spatial-semantic}
        \mathcal{L}_{d_s}(I_1,I_2,I_0)&=\frac{1}{M}\sum_{i=1}^M{\frac{\mathcal{L}_c(d^i_1, d^{i +}_{2\&0,s}, \mathbb{D}^{i -}_{2,s})}{(\sigma_1^i)^{-1}}+\log{(\sigma_1^i)^{-1}}} \text{,}
    \end{align}
    \begin{align}
        \label{formula:infonce_spatial-category}
        \mathcal{L}_{d_c}(I_1,I_2,I_0)&=\frac{1}{M}\sum_{i=1}^M{\mathcal{L}_c(d^i_1, d^{i +}_{2\&0,c}, \mathbb{D}^{i -}_{2,c})} \text{,}
    \end{align}  
    where $d^{i +}_{2\&0,s}$ denotes the union of the positive samples $d^{i +}_{2,s}$ from $I_2$ and $d^{i +}_{0,s}$ from $I_0$ for the $i$-th query, and $d^{i +}_{2\&0,c}$ is defined similarly.

	\section{Experiments}\label{sec:exp}
	
	\noindent
    \textbf{Training data generation.}
    To bootstrap the object-centric keypoint detection and description, we first create a large-scale object-clustered synthetic dataset that consists of 21 objects from the YCB-Video dataset~\cite{posecnn}. Furthermore, to align the synthetic and real domains, we refer to the idea of physically plausible domain randomization (PPDR)~\cite{se3} to generate the scenes where objects can be fallen onto the table/ground with preserving physical properties. The viewpoint of the camera is randomly sampled from the upper hemisphere. We construct a total of 220 scenes where each contains 6 objects and acquire a continuous sequence of images from each scene, resulting in 33k images. More examples of the dataset can be found in the appendix.
    
    \noindent
    \textbf{Training.} 
    We choose 20 keypoints of each object to construct positive-negative pairs, and the temperature $\tau$ in intra-object InfoNCE~\citep{infonce} loss and inter-object InfoNCE loss are set to 0.07, 0.2 respectively. 
    The data augmentation is composed of color jittering, random gray-scale conversion, gaussian noise, gaussian blur, and random rotation. The $\delta$ and $N$ are set to 8 and 16 pixels. We set the trade-off weights of two subparts of descriptor $\lambda_1=1$ and $\lambda_2=1$. 
    
    Our model is implemented in PyTorch~\citep{pytorch} with a mini-batch size of 4 and optimized with the Adam~\citep{kingma2017adam} for 20 epochs, and all the input images are cropped to $320 \times 320$. We use a learning rate of $10^{-4}$ for the first 15 epochs, which is dropped ten times for the remainder. 
    
    \noindent
    \textbf{Testing.} To reduce the domain discrepancy between synthetic and real data, we modify the statistics of BN~\citep{bn} layers learned in simulation for adapting the model to real scenes. Strictly speaking, we suppose that the real test data cannot be accessed, so we only use the mean and variance of BN layers from a current real image, i.e., the batch size is set to 1, and do not update the statistics. For comparative evaluations, we record the best results of each baseline which adopts the BN layer, with or without this trick.

    \noindent
    \textbf{Baselines.} 
    As a classical keypoint detection and description, we choose the handcrafted method SIFT~\cite{sift} as the baseline. We also compare against Superpoint~\cite{superpoint}, R2D2~\cite{r2d2} and DISK~\cite{disk} which are data-driven methods.
	
	\begin{table*}[htbp]
      \centering
      \caption{Quantitative evaluation for real-real image matching. }
      \resizebox{0.97\textwidth}{!}{
        \begin{tabular}{l|ccc|ccc|ccc|ccc|ccc}
    \multirow{2}[1]{*}{Objetcs} & \multicolumn{3}{c|}{SIFT(128)} & \multicolumn{3}{c|}{Superpoint(256)} & \multicolumn{3}{c|}{R2D2(128)} & \multicolumn{3}{c|}{DISK(128)} & \multicolumn{3}{c}{Ours(96)} \\
          & Kpts  & MMA5  & MMA7  & Kpts  & MMA5  & MMA7  & Kpts  & MMA5  & MMA7  & Kpts  & MMA5  & MMA7  & Kpts  & MMA5  & MMA7 \\
    \midrule
    cracker\_box & 72.1  & 23.4\% & 27.4\% & 21.9  & 25.8\% & 32.7\% & 26.1  & 22.6\% & 29.0\% & 41.2  & 26.2\% & 32.1\% & \textbf{122.2 } & \textbf{37.8\%} & \textbf{49.8\%} \\
    sugar\_box & 25.2  & 5.6\% & 6.8\% & 9.2   & \textbf{19.1\%} & 25.3\% & 12.8  & 9.6\% & 15.0\% & 14.4  & 15.7\% & 21.0\% & \textbf{64.4 } & 18.9\% & \textbf{29.4\%} \\
    tomato\_soup\_can & 21.3  & 8.7\% & 11.6\% & 9.7   & 47.8\% & 57.1\% & 9.6   & 41.3\% & 46.0\% & 11.7  & \textbf{66.8\%} & \textbf{73.4\%} & \textbf{54.0 } & 60.1\% & 70.1\% \\
    mustard\_bottle & 27.6  & 18.8\% & 21.6\% & 10.2  & 25.6\% & 32.5\% & 12.8  & 26.3\% & 29.9\% & 19.2  & 41.2\% & 51.9\% & \textbf{88.1 } & \textbf{43.6\%} & \textbf{61.1\%} \\
    bleach\_s & 27.9  & 11.6\% & 12.0\% & 10.7  & 27.6\% & 35.0\% & 16.6  & 19.8\% & 22.9\% & 15.9  & 22.8\% & 25.4\% & \textbf{72.7 } & \textbf{33.6\%} & \textbf{42.2\%} \\
    \midrule
    ALL   & 34.8  & 13.6\% & 15.9\% & 12.3  & 29.2\% & 36.5\% & 15.6  & 23.9\% & 28.5\% & 20.5  & 34.5\% & 40.7\% & \textbf{80.3 } & \textbf{38.8\%} & \textbf{50.5\%} \\
    \end{tabular}}%
    \label{tab:real_real}%
    \end{table*}%

	\subsection{Image matching}\label{sec:image_matching}

    All the methods are evaluated on the following datasets.
    
	\textit{YCB-Video}~\cite{posecnn} consists of 21 objects and 92 RGB-D video sequences with pose annotations. We use the 2,949 keyframes in 12 videos which are commonly evaluated in other works. In this scene, all the objects are static, and the camera is moving with a slight pose change.
	
	\textit{YCBInEOAT~\cite{se3}} consists of 9 video sequences and each video has one manipulated object from YCB-Video. In this dataset, objects are translated and rotated by different end-effectors while the camera is static. We selected 5 valid videos with a total of 1112 keyframes.

    We set two object-centric image matching tasks. 
    \begin{itemize}
        \item \textbf{Synthetic-real matching.} The test images are 2949 keyframes from YCB-Video. Two adjacent frames are selected where the next frame is adopted as a target (real) image, and the rendered (synthetic) images on the previous pose of each object are used as the references. Pairs are matched and filtered by RANSAC~\cite{ransac}. 
    	\item \textbf{Real-real matching.} The keypoints and descriptors from the manipulated object whose mask is known in the initial frame of each video, predicted by keypoint methods, are matched with the subsequent frames (targets) to show the tracking performance on the object keypoints. 
	\end{itemize}
	In this work, the bounding box or mask of each object in the target frame is not provided. We utilize the nearest neighbor search to find the matched keypoints from two views, i.e., mutual nearest neighbors are considered matches. We adopt the Mean Matching Accuracy (MMA)~\cite{mma} for matching evaluation, i.e., the average percentage of correct matches per image pair. A correct match represents its reprojection error, which is below a given matching threshold. We record the MMA5 and MMA7 of each object with an error threshold of 5 and 7 pixels, respectively. Each method would detect top-5k keypoints per image. Kpts means the average number of matches per object per image. 
    
\begin{figure}[!ht]
        \centering
        \includegraphics[width=0.45\textwidth]{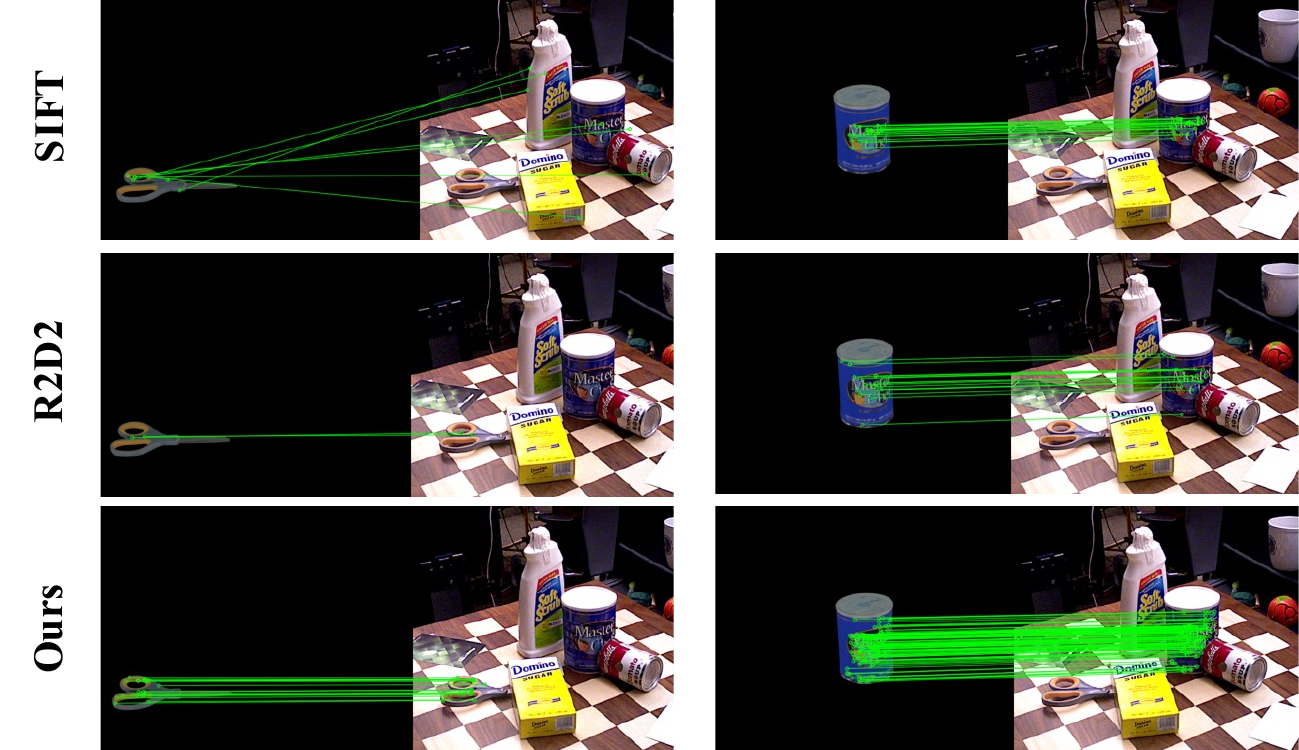}
        \caption{Synthetic-real image matching.}
        \label{fig:syn-real}
    \end{figure}

    \noindent
    \textbf{Comparison to baselines.}
	In terms of synthetic-real matching, our object-centric method significantly outperforms the scene-centric methods, as shown in Table~\ref{tab:syn_real}. Our method surpasses all the counterparts by a large margin, i.e., more than $20\%$ in the MMA5 and MMA7.
	We can extract more matching keypoints with broader distribution on the surface of the objects, while the matches are not affected by other occluded objects, as shown in Figure.~\ref{fig:syn-real}. It should be noted that the scene-level methods can not extract matching keypoints in the junction of object and background, since the different backgrounds between the target and the rendered objects.

	\begin{table}[!ht]
      \centering
      \vskip  -0.1in
      \caption{Quantitative evaluation for synthetic-real image matching. The last three metrics is averaged of each object. Dim=length of descriptors.}
      \vskip-0.1in
        \resizebox{0.38\textwidth}{!}{
        \begin{tabular}{lcccc}
        Method & Dim   & Kpts  & MMA5 & MMA7 \\
        \midrule
        SIFT  & 128   & 16.9  & 24.2\%  & 30.1\%  \\
        Superpoint & 256   & 15.7  & 33.6\% & 43.8\% \\
        R2D2  & 128   & 20.0  & 34.8\% & 44.6\% \\
        DISK  & 128   & 15.8  & 28.2\% & 35.1\% \\
        \midrule
        Ours  & 96 & \textbf{92.1} & \textbf{50.0\%} & \textbf{57.2\%} \\
        \end{tabular}}%
      \label{tab:syn_real}%
    \end{table}%
    
    
\begin{figure}[!ht]
        \centering
        \includegraphics[width=0.45\textwidth]{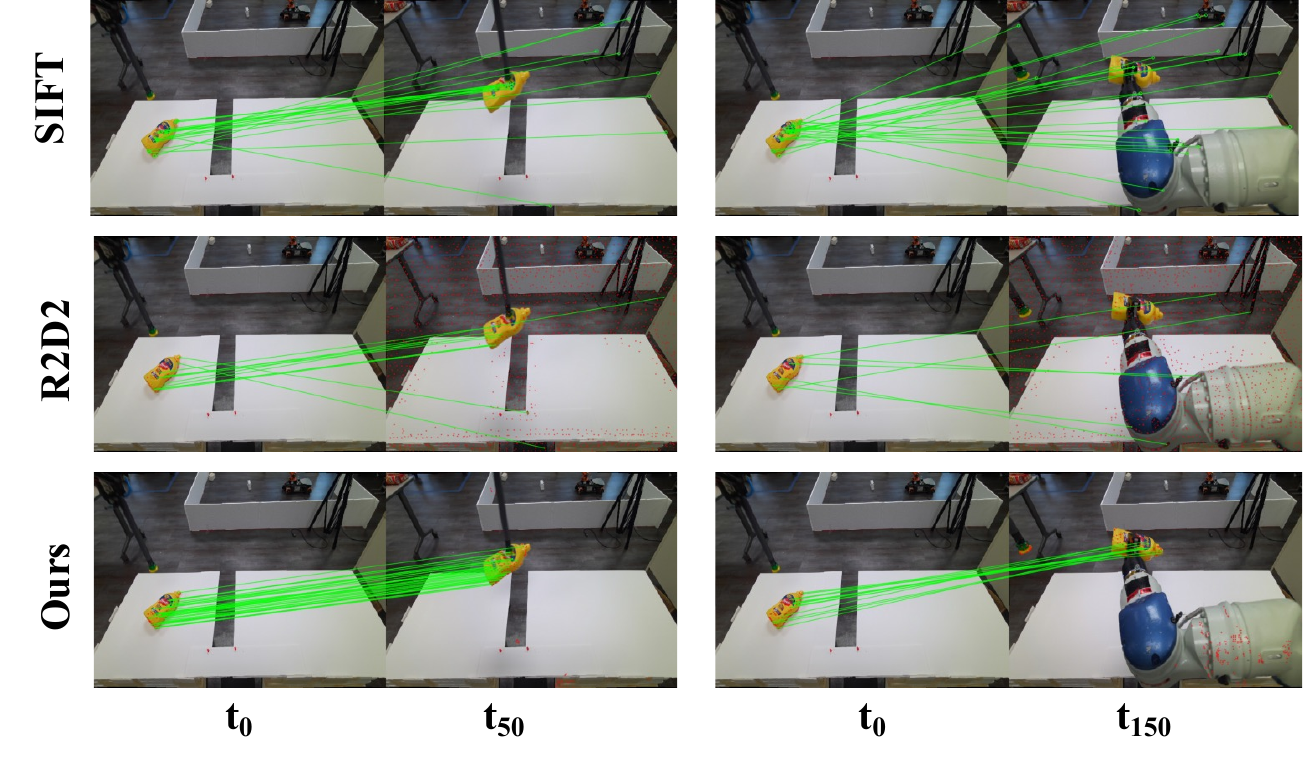}
        \caption{Real-real image matching. $t_k$ = $k$-th frame.}
        \label{fig:real_real}
    \end{figure}

	In Table~\ref{tab:real_real}, we further provide a quantitative comparison with the baselines in the real-real matching track.
	Compared with the baselines, our method also clearly attains consistent improvements. And it shows that our method outperforms the Superpoint by at least $68.0$ Kpts, the R2D2 by $64.7$ Kpts and the DISK $59.8$ Kpts. In the MMA7, our method surpasses the DISK by $9.8\%$ and the Superpoint by $14.0\%$. Despite the obvious movement of the target in the scene, our method can still detect the matching points. In contrast, the performance of other methods will deteriorate significantly, as shown in Figure.~\ref{fig:real_real}. It shows the encouraging generalization ability of our method from simulation to reality. More details of matching results can be seen in the appendix.
	
    \subsection{6D pose estimation}\label{sec:6d_pose}
    \begin{table*}[htbp]
      \centering
      \caption{Evaluation for 6D pose estimation on YCB-Video dataset.}
        \resizebox{0.93\textwidth}{!}{
        \begin{tabular}{l|cc|cc|cc|cc|cc|cc}
     & \multicolumn{10}{c|}{Keypoint-Based Method}                                   & \multicolumn{2}{c}{End2End Method} \\
\cmidrule{1-11}          & \multicolumn{8}{c|}{Scene-Centric}                            & \multicolumn{2}{c|}{Object-Centric} & \multicolumn{2}{c}{(Supervised)} \\
    \midrule
    \multirow{2}[4]{*}{Objects} & \multicolumn{2}{c|}{SIFT(128)} & \multicolumn{2}{c|}{S-Point(256)} & \multicolumn{2}{c|}{R2D2(128)} & \multicolumn{2}{c|}{DISK(128)} & \multicolumn{2}{c|}{Ours(96)} & \multicolumn{2}{c}{PoseCNN} \\
\cmidrule{2-13}          & ADD   & ADDS  & ADD   & ADDS  & ADD   & ADDS  & ADD   & ADDS  & ADD   & ADDS  & ADD   & ADDS \\
    \midrule
    master\_chef\_can & 20.1  & 38.4  & \textbf{48.5} & 76.8  & 31.6  & 56.2  & 22.2  & 44.9  & 48.2  & \textbf{77.5} & 50.9  & 84.0  \\
    cracker\_box & 0.0   & 3.6   & \textbf{55.2} & \textbf{69.0} & 39.8  & 57.9  & 46.5  & 60.1  & 35.4  & 60.3  & 51.7  & 76.9  \\
    sugar\_box & 28.7  & 35.8  & 65.1  & 79.0  & 53.0  & 61.5  & 49.2  & 60.2  & \textbf{74.6} & \textbf{86.3} & 68.6  & 84.3  \\
    tomato\_soup\_can & 19.4  & 27.0  & 42.1  & 57.9  & 49.1  & 60.8  & 33.1  & 42.6  & \textbf{56.5} & \textbf{75.3} & 66.0  & 80.9  \\
    mustard\_bottle & 9.7   & 13.1  & 47.1  & 52.2  & 40.4  & 46.8  & 31.0  & 38.3  & \textbf{54.5} & \textbf{72.8} & 79.9  & 90.2  \\
    tuna\_fish\_can & 7.8   & 11.4  & 13.8  & 21.1  & 1.6   & 2.2   & 0.2   & 0.7   & \textbf{53.6} & \textbf{73.5} & 70.4  & 87.9  \\
    pudding\_box & 2.9   & 6.8   & 5.8   & 8.2   & 0.5   & 0.7   & 5.3   & 8.8   & \textbf{37.9} & \textbf{49.0} & 62.9  & 79.0  \\
    gelatin\_box & \textbf{68.2} & \textbf{79.8} & 55.4  & 68.0  & 34.1  & 39.4  & 54.9  & 65.0  & 55.7  & 70.8  & 75.2  & 87.1  \\
    potted\_meat\_can & 8.1   & 11.4  & 30.4  & 38.4  & 23.3  & 32.1  & 24.1  & 30.4  & \textbf{51.9} & \textbf{70.5} & 59.6  & 78.5  \\
    banana & 0.3   & 0.6   & 0.0   & 0.8   & 0.3   & 0.5   & 0.5   & 2.2   & \textbf{11.5} & \textbf{28.5} & 72.3  & 85.9  \\
    pitcher\_base & 0.0   & 3.2   & 3.4   & 9.8   & 0.9   & 3.2   & 0.2   & 3.8   & \textbf{16.9} & \textbf{30.9} & 52.5  & 76.8  \\
    bleach\_cleanser & 17.6  & 22.4  & \textbf{43.3} & \textbf{54.4} & 37.2  & 49.5  & 38.7  & 50.3  & 39.8  & 55.2  & 50.5  & 71.9  \\
    bowl  & 0.0   & 0.3   & 0.2   & 0.9   & 0.2   & 0.7   & 0.3   & 3.5   & \textbf{3.7} & \textbf{26.8} & 6.5   & 69.7  \\
    mug   & 0.2   & 0.3   & 0.2   & 0.9   & 0.2   & 0.2   & 0.3   & 1.2   & \textbf{14.3} & \textbf{45.7} & 57.7  & 78.0  \\
    power\_drill & 0.9   & 2.9   & 55.8  & 63.3  & 36.5  & 43.1  & 15.2  & 21.5  & \textbf{61.4} & \textbf{76.5} & 55.1  & 72.8  \\
    wood\_block & 0.0   & 2.2   & 0.8   & 4.1   & 0.0   & 0.0   & 0.0   & 0.4   & \textbf{3.2} & \textbf{22.1} & 31.8  & 65.8  \\
    scissors & 0.0   & 0.0   & 1.0   & 2.1   & 0.0   & 0.0   & 0.0   & 0.6   & \textbf{20.9} & \textbf{38.3} & 35.8  & 56.2  \\
    large\_marker & 22.3  & 26.9  & 24.3  & 34.8  & 17.2  & 18.9  & 0.6   & 0.7   & \textbf{56.8} & \textbf{68.5} & 58.0  & 71.4  \\
    large\_clamp & 0.0   & 0.4   & 1.0   & 3.8   & 0.0   & 0.1   & 0.1   & 0.5   & \textbf{14.8} & \textbf{36.8} & 25.0  & 49.9  \\
    ex\_large\_clamp & 0.0   & 0.6   & 0.6   & 4.2   & 0.1   & 0.5   & 0.2   & 0.5   & \textbf{14.5} & \textbf{45.5} & 15.8  & 47.0  \\
    foam\_brick & 0.0   & 0.0   & 0.7   & 1.3   & 0.0   & 0.0   & 0.6   & 1.2   & \textbf{43.4} & \textbf{69.5} & 40.4  & 87.8  \\
    \midrule
    ALL   & 10.6  & 15.2  & 30.3  & 39.9  & 23.4  & 30.6  & 19.0  & 25.9  & \textbf{42.2} & \textbf{61.9} & 53.7  & 75.9  \\
    \end{tabular}
    
 }%
      \label{tab:6d_pose}%
    \end{table*}%
    
    \noindent
    \textbf{Evaluation protocol.}
    Pipeline of pose estimation : (1) Render multiple images in different poses of the test objects as templates. To balance the evaluation speed and accuracy, we rendered 96 templates for each object. (2) Match the templates with the real images one by one, and select the best template according to the number of matched pairs. (3) 6D pose can be solved through the Perspective-n-Point (PnP) and RANSAC algorithms.
    We report the average recall ($\%$) of ADD(-S) for pose evaluation which is the same as in PoseCNN~\cite{posecnn}.
    
\begin{figure}[!ht]
        \centering
        \includegraphics[width=0.47\textwidth]{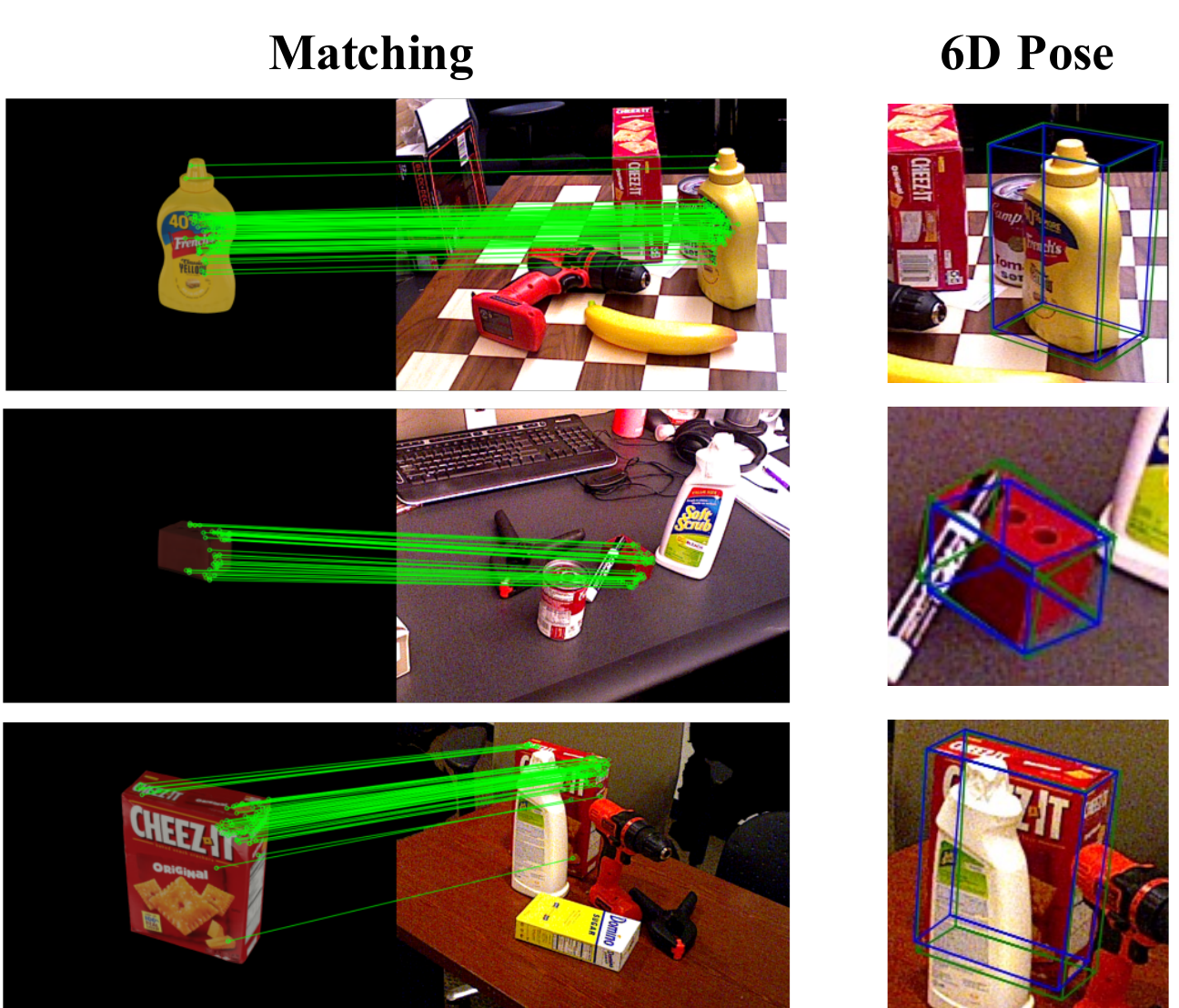}
        \caption{Examples of 6D object pose estimation results on YCB-Video by our method. Left: the keypoints matching between the best template and a real image. Right: the pose estimation results, where the green bounding boxes stands for ground truths while the blue boxes are our predictions.}
        \label{fig:6d_pose}
    \end{figure}

    
    \noindent
    \textbf{Comparison to baselines.}
    In Table~\ref{tab:6d_pose}, our method achieves the state-of-the-art performance among other keypoint-based methods with a large margin (approximately $20\%$ on both ADD and ADD-S). The pose estimation results of each object can be seen in the appendix. By introducing an object-centric mechanism, our method can significantly bootstrap performance on 6D pose estimation. As shown in Figure~\ref{fig:6d_pose}, the keypoint matching result indicates our method can serve a crucial role on downstream 6D pose estimation tasks, even in the occluded and clustered environment.
    Figure.~\ref{fig:6d_pose} displays some image matching and 6D pose estimation results. It can be seen that ours can detect matching points even when the object is severely occluded or in a large pose difference between a template and target. 
    
    \noindent
    \textbf{Comparison to un/weak-supervised pose estimation.}
    Self6D~\cite{self6d} is a sim2real pose estimation method, where the model is fully trained on the synthetic RGB data in a self-supervised way. Further, the model is fine-tuned on an unannotated real RGB-D dataset, called self6D(R). Note that the self6D(R) is a weak-supervised baseline due to the access to real-world data. In Table~\ref{tab:un_6d_pose}, our method achieves an overall average recall of 59.4\%, which surpasses 10.8\% of Self6d(R) and 29.5\% of Self6D. And our method brings the encouraging generalization ability from simulation to reality, while the performance of PoseCNN only trained in the simulation would drop dramatically.
    
    \begin{table}[!ht]
      \centering
      \caption{Results for un/weak-supervised on YCB-Video dataset.  W-Supervised=Weak-Supervised.}
      \resizebox{0.45\textwidth}{!}{
        \begin{tabular}{l|c|c|c|c}
              & \small{W-Supervised} & \multicolumn{3}{c}{\small{Sim2Real/Unsupervised}} \\
        \midrule
           \small{Objects}  & \small{Self6D(R)} & \small{Self6D} & \multicolumn{1}{l|}{\small{PoseCNN}} & Ours \\
        \midrule
        \small{mustard\_bottle} & \textbf{88.2}  & 73.7  & 3.7   & 72.8  \\
        \small{tuna\_fish\_can} & 69.7  & 26.6  & 3.1   & \textbf{73.5}  \\
        \small{banana} & 10.3  & 4.0   & 0.0   & \textbf{28.5}  \\
        \small{mug}   & 43.4  & 23.9  & 0.0   & \textbf{45.7}  \\
        \small{power\_drill} & 31.4  & 21.4  & 0.0   & \textbf{76.5}  \\
        \midrule
        ALL   & 48.6  & 29.9  & 1.4   & \textbf{59.4}  \\
        \end{tabular}}%
      \label{tab:un_6d_pose}%
    \end{table}%

	\subsection{Ablation study}\label{sec:ablation_study}
	In this subsection, we will explore the sensitivity of our method in terms of reliability threshold and perform a diverse set of analyses on assessing the impact of each component that contributes to our method. 
    
    \begin{table}[!ht]
      \centering
      \caption{Sensitivity to reliability threshold $r_{thr}$. }
        \resizebox{0.46\textwidth}{!}{
        \begin{tabular}{c|cc|cc|cc}
        \multirow{2}[1]{*}{$r_{thr}$} & \multicolumn{2}{c|}{Syn-Real} & \multicolumn{2}{c|}{Real-Real} & \multicolumn{2}{c}{6D Pose} \\
              & Kpts  & MMA5  & Kpts  & MMA5  & ADD   & ADDS \\
        \midrule
        0.0   & 10.0  & 14.2\% & 43.7  & 3.0\% & 15.8  & 24.3  \\
        1.5   & 92.1  & 50.0\% & 80.3  & 38.8\% & 42.2  & 61.9  \\
        3.0   & 80.4  & 45.1\% & 78.9  & 39.3\% & 40.2  & 59.3  \\
        4.5   & 71.1  & 41.0\% & 67.7  & 41.0\% & 36.1  & 53.3  \\
        \end{tabular}}%
      \label{tab:ablation_kp}%
    \end{table}%
    
	\noindent
    \textbf{Sensitivity of confidence threshold.}
    Table~\ref{tab:ablation_kp} illustrates the performance of keypoint matching and pose estimation under various confidence threshold $r_{thr}$, i.e, 0.0, 1.5, 3.0, 4.5. In particular, $r_{thr}$ = 0 means that all pixels may be selected as keypoints. Obviously, such performance will be significantly deteriorated. We use confidence threshold $r_{thr}$ being 1.5 as the default setting, as the performance becomes superior in all tasks.
    We visualize the dense detector heatmap of some objects, i.e, the confidence of pixels. 
    

    
    
    \begin{figure}[htbp]
        \centering
        \includegraphics[width=0.45\textwidth]{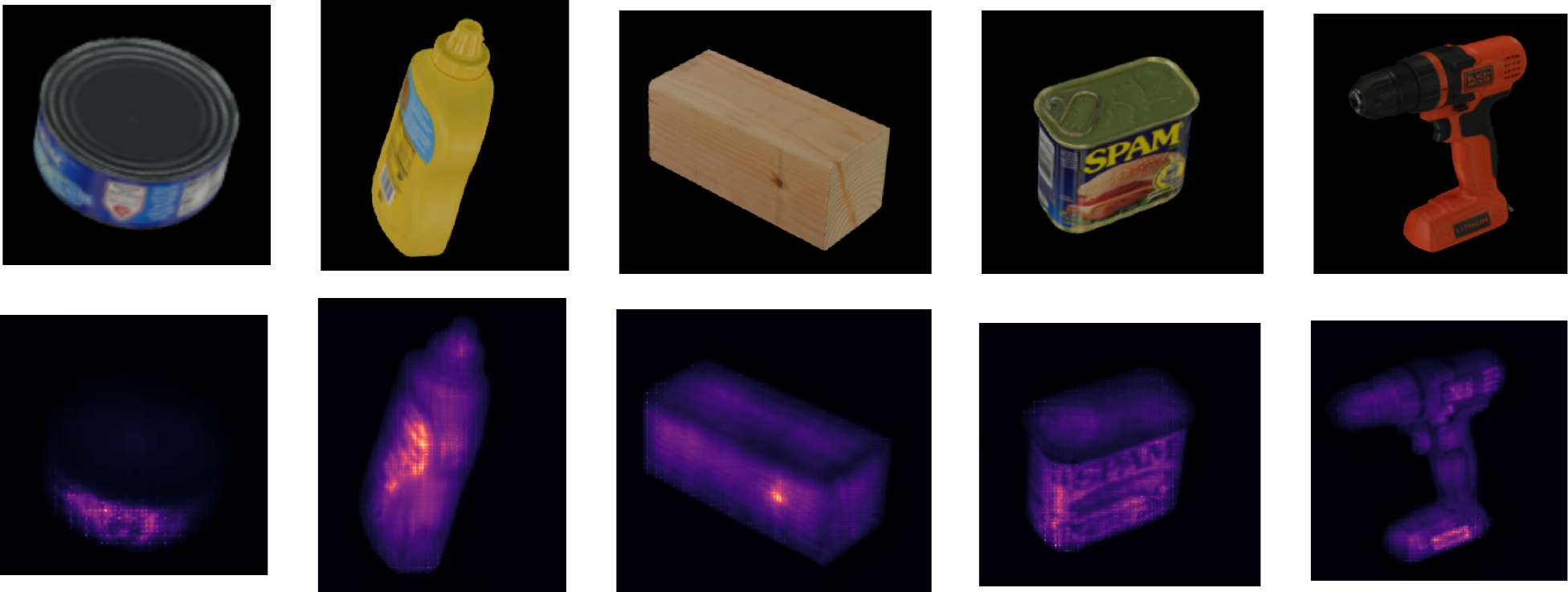}
        \caption{Visualization of confidence heatmap in the detector. The highlighted regions are selected as keypoints.}
        \label{fig:keypoint_vis}
    \end{figure}

    \noindent
    \textbf{Training Strategy.} 
    In Table~\ref{tab:ablation_des}, we compare different training strategies used to learn detector and descriptor. 1) if we randomly select negative samples from all the pixels, like R2D2, it will exhibit extremely poor performance. Object-centric sample strategy will be essential for object keypoint detection. 2) Each module, including the repeatability of keypoint detector ($\mathcal{L}_r$), the decoupled descriptor (Dec.), and the semantics consistency (Sem.) have positive effects on the final result.

    \begin{table}[htbp]
      \centering
      \vskip -0.1in
      \caption{Ablation study of training strategy. NEG./R.= negatives randomly sampled from all the pixels; NEG./Obj.= object-centric sampling; $\mathcal{L}_{r}$= repeatability of detector; Dec.= decoupled descriptors; Sem.= semantics consistency. We \textbf{bold} the best and \underline{underline} the second best.}
      \resizebox{0.46\textwidth}{!}{
        \begin{tabular}{c|rcccc|cccc}
        \multirow{2}[1]{*}{Strategy} & \multicolumn{2}{c}{NEG} & \multirow{2}[1]{*}{$\mathcal{L}_{r}$} & \multirow{2}[1]{*}{Dec.} & \multicolumn{1}{c|}{\multirow{2}[1]{*}{Sem. }} & Syn-Real & Real-Real & \multicolumn{2}{c}{6D Pose} \\
              & \multicolumn{1}{c}{R.} & Obj.  &       &       &       & MMA5  & MMA5  & ADD   & ADDS \\
        \midrule
        r2d2-like & \multicolumn{1}{c}{\checkmark} &       &       &       &       & 1.1\%  & 2.3\%  & 1.5  & 2.0  \\
              &       & \checkmark   &       &       &       & 46.9\% & 29.3\% & 22.1  & 33.8 \\
              &     & \checkmark   & \checkmark   &       &       & 45.5\% & 35.1\% & 26.7  & 42.7  \\
              &       & \checkmark   & \checkmark   & \checkmark   &       & 47.8\% & \textbf{44.4\%} & \underline{40.8}  & \underline{58.2}  \\
              &       & \checkmark   & \checkmark   &       & \checkmark   & \textbf{52.4\%} & 30.0\% & 37.6  & 54.7  \\
        Ours  &       & \checkmark   & \checkmark   & \checkmark   & \checkmark   & \underline{50.0\%} & \underline{38.8\%} & \textbf{42.2} & \textbf{61.9} \\
        
        \end{tabular}}%
        \vskip -0.1in
      \label{tab:ablation_des}%
    \end{table}%

    \subsection{Model efficiency and generalization}\label{sec:generalization}
    \textbf{Efficiency.} 
    Our model costs about 0.51s to extract keypoints and descriptors from a $640 \times 480$ image, while SIFT, R2D2, Superpoint, and DISK take about 0.04s, 0.10s, 0.19s, and 0.48s, respectively. It indicates that the computation overhead by our method is acceptable, particularly given the remarkable improvement in performance. \\
    \textbf{Generalization. }
    The generalization ability of our method lies in two aspects. The first one is the sim2real adaptation. Our model is trained on simulation data and tested straightly on real images. Such sim2real generalization benefits the field of robotic perception/manipulation where the keypoint annotations are difficult to obtain, but the CAD model of a target object is always given. 
    The second one is the generalization to unseen objects. To illustrate, we provide the matching evaluations on objects with simulated images outside the training set in Table~\ref{tab:generalization}. Twenty unseen objects are selected from the OCRTOC dataset~\cite{liu2021ocrtoc}, in which half of them have the same class as YCB-Video objects but with different shapes or textures (seen class), and other objects with novel class have not been seen in training (unseen class). The evaluation protocol is similar to real-real matching in~\ref{sec:image_matching}. From Table~\ref{tab:generalization}, it can be seen that our model also achieves satisfactory matching performance for unseen objects. More details are provided in the appendix. \\

    \begin{table}[ht]
      \centering
      \caption{Image matching evaluation on unseen objects.}
      \vskip-0.1in
        \begin{tabular}{c|cc|cc}
        \small{Method} & \multicolumn{2}{c|}{\small{Seen class} } & \multicolumn{2}{c}{\small{Unseen class} } \\
              & \small{Kpts}  & \small{MMA5}  & \small{Kpts}  & \small{MMA5} \\
        \midrule
        SIFT  & 23.3  & 32.5\% & 16.8  & 29.6\% \\
        R2D2  & 21.2  & 61.3\% & 16.1   & 49.3\% \\
        Ours  & \textbf{90.5} & \textbf{65.4\%} & \textbf{75.9} & \textbf{61.3\%} \\
        \end{tabular}%
      \label{tab:generalization}%
    \end{table}%

	\section{Conclusion}\label{sec:conclusion}
	We present for the first time a sim2real contrastive learning framework for object-centric keypoint detection and description, which is only trained from synthetic data. Our experiments demonstrate that (1) our object keypoint detector and descriptor can be robust for both synthetic-to-real and real-to-real image matching tasks, and (2) our method leads to a superior result on unsupervised (sim2real) 6D pose estimation. Future work may explore integrating 2D and 3D inputs to find more repeatable keypoints and distinctive descriptors for texture-less objects. 
	

\section*{Acknowledgment}
This research was funded by the National Science and Technology Major Project of the Ministry of Science and Technology of China (No.2018AAA0102900).  Meanwhile, this work is jointly sponsored by he National Natural Science Foundation of China (Grant No. 62006137) and CAAI-Huawei MindSpore Open Fund. It was also partially supported by the National Science Foundation of China (NSFC) and the German Research Foundation (DFG) in the project Cross Modal Learning, NSFC 61621136008/DFG TRR-169.
	
	{\fontsize{9pt}{10pt} \selectfont 
		\bibliography{main}
	}
	
	\clearpage
	\begin{appendix}
	\section{Appdendix}
	\subsection{Dataset Visualization}
\begin{figure*}
    \centering
    \subfigure[Sequence 1]{
    \begin{minipage}[b]{0.95\textwidth}
    \includegraphics[width=1\textwidth]{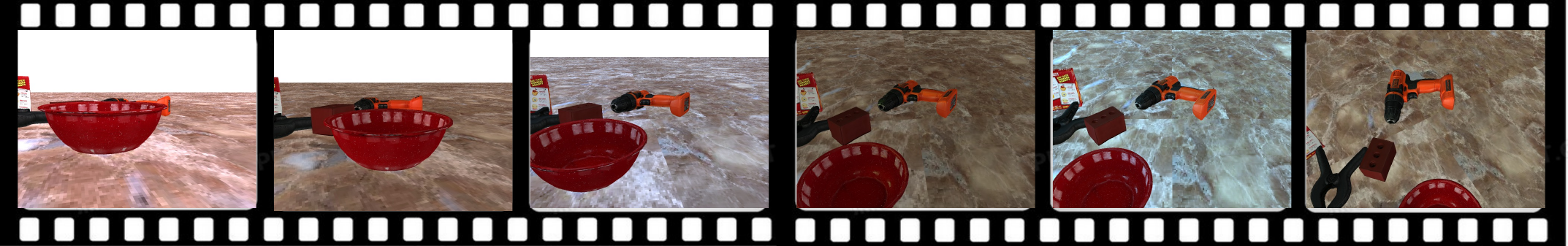}
    \end{minipage}}
    \subfigure[Sequence 2]{
    \begin{minipage}[b]{0.95\textwidth}
    \includegraphics[width=1\textwidth]{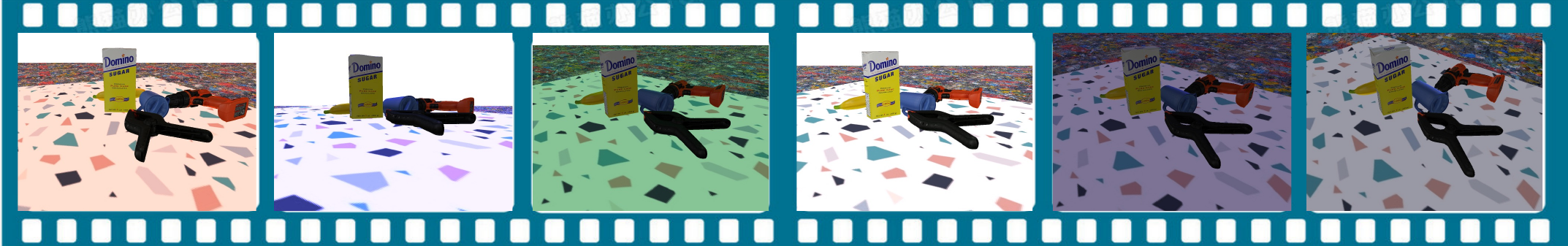}
    \end{minipage}}
    \caption{Visualization of some sequence images generated from Pybullet simulator. The parameters of the light color, light direction, light distance and the viewpoint of the camera are randomly sampled in each frame.} 
    \label{fig:data_generation}
\end{figure*}

We generate the scenes where objects can be fallen onto the table/ground with preserving physical properties. The color of the background and the illumination are randomized. We construct a total of 220 scenes where each scene contains 6 objects and acquire a continuous sequence of images for each scene, resulting in 33k synthetic images. Part of the scenes in the dataset are shown in Figure~\ref{fig:data_generation}.

\subsection{Network Architecture}
For the encoder, we apply Unet~\cite{unet} plus the first three ResNet18~\cite{resnet} blocks and three upsampling layers as the backbone. For the keypoint detector, the element-wise square operation and $1\times1$ convolution layer are applied. And for the keypoint descriptor, we use two decoupled descriptors with a $\ell_2$ normalization. The architectures of the U-net decoder, keypoint detector and descriptor are shown in Table~\ref{tab:network}. There are some notations: \textbf{ker} is the kernel size; \textbf{stri} is the side size; 
\textbf{chan} is the dimensionality of output channels; \textbf{res} is the downscaling factor for each layer; \textbf{input} indicates the input of each layer where ↑ is a 2× nearest-neighbor upsampling of the layer; $x^2$ is an element-wise square and $\ell_2$-norm is a L2 normalization.

\begin{table}[htbp]
  \centering
   \caption{\textbf{Network architecture.} }
   \vspace{5pt}
  \resizebox{0.98\columnwidth}{!}{
  \begin{tabular}[t]{l}

\begin{tabular}[t]{l|l|l|l|l|l|l}
\toprule[ 2pt]
\multicolumn{7}{l}{\textbf{Unet-Decoder}} \\
\hline
\textbf{layer} & \textbf{ker} & \textbf{stri} & \textbf{chan} & \textbf{res} & \textbf{input}   & \textbf{activation}    \\ \hline
upconv4       & 3      & 1      & 256      & 16    & econv4                    & ELU \\
iconv4        & 3      & 1      & 256      & 8     & $\uparrow$upconv4, econv3 & ELU \\ \hline

upconv3       & 3      & 1      & 128       & 8     & iconv4                    & ELU \\
iconv3        & 3      & 1      & 128       & 4     & $\uparrow$upconv3, econv2 & ELU \\ \hline

upconv2       & 3      & 1      & 96       & 4     & iconv3                    & ELU \\
iconv2        & 3      & 1      & 96       & 2     & $\uparrow$upconv2, econv1 & ELU \\ \hline

upconv1       & 3      & 1      & 96       & 2     & iconv2                    & ELU \\
iconv1        & 3      & 1      & 96       & 1     & $\uparrow$upconv1         & - \\  \hline

\end{tabular}  \\

\begin{tabular}[t]{l|l|l|l|l|l|l}
\toprule[ 2pt]
\multicolumn{7}{l}{\textbf{Keypoint Detector}} \\
\hline
\textbf{layer} & \textbf{ker} & \textbf{stri} & \textbf{chan} & \textbf{res} & \textbf{input}   & \textbf{activation}    \\ \hline
square0         & -      & -      & 96      & 1     & iconv1                    & $x^2$ \\ \hline
kconv0         & 1      & 1      & 1      & 1     & square0                    & - \\ \hline
\end{tabular} \\

\begin{tabular}[t]{l|l|l|l|l|l|l}
\toprule[ 2pt]
\multicolumn{7}{l}{\textbf{Keypoint Descriptor}} \\
\hline
\textbf{layer} & \textbf{ker} & \textbf{stri} & \textbf{chan} & \textbf{res} & \textbf{input}   & \textbf{activation}    \\ \hline
intra-obj0         & -      & -      & 64      & 1     & iconv1[:, :64]                    & $\ell_2$-norm \\ \hline
inter-obj0         & -      & -      & 32      & 1     & iconv1[:, 64:]                    & $\ell_2$-norm \\ \hline

\end{tabular}
\end{tabular} 
 }
\label{tab:network}
\end{table}

\subsection{Additional Results}
\subsubsection{Synthetic-real image matching}

\begin{figure*}[!ht]
    \centering
    \includegraphics[width=0.8\textwidth]{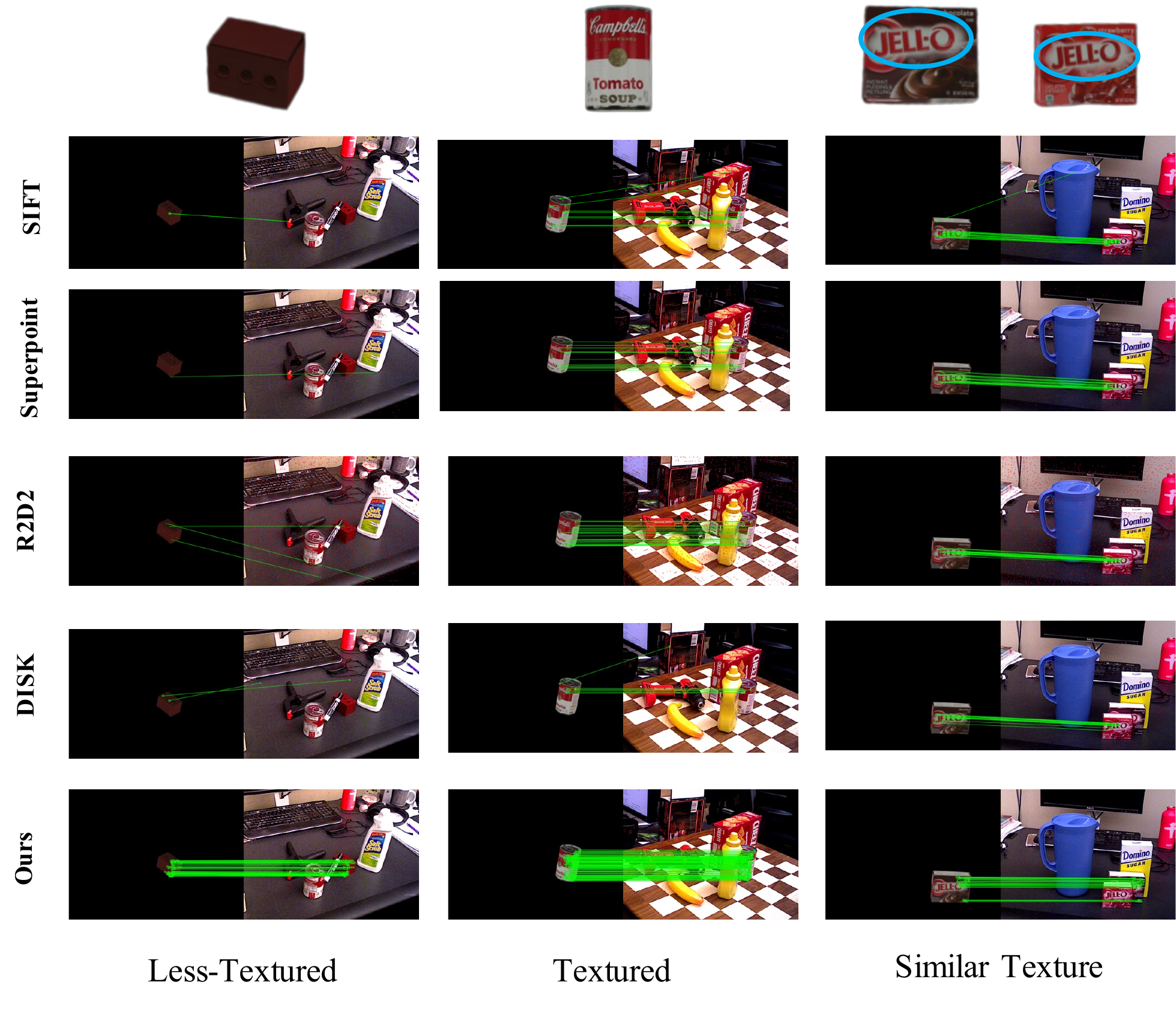}
    \vskip -0.1in
    \caption{Visualization of synthetic-real image matching on some typical objects.}
    \label{fig:syn_real}
\end{figure*}

The matching results of each object can be seen in Table~\ref{tab:syn-real}.
Our object-centric method significantly surpasses the scene-centric techniques by a large margin on average. Especially for the less-textured objects, our method can extract more keypoints and have robust matching performance. 
As shown in Figure~\ref{fig:syn_real}, although scene-centric methods can find some matches in terms of rich textured objects, we extract more matching keypoints with broader distribution on the surface of the objects. For some challenging objects with similar textures, such as the same word 'JELLO' in two CAD object models, scene-centric methods are unanimously cheated by the local similarity and predict wrong matches. Thanks to the object-wise discrimination, our method accurately predicts the object correspondence and matches the keypoints on the same object.

\begin{table*}[htbp]
  \centering
  \caption{Quantitative evaluation for synthetic-real image matching. Less-textured objects are underlined. Kps=number of matched keypoints per object per image.}
  \vspace{8pt}
  \resizebox{0.95\textwidth}{!}{
    \begin{tabular}{l|cc|cc|cc|cc|cc}
    \multirow{2}[1]{*}{Objects} & \multicolumn{2}{c|}{SIFT} & \multicolumn{2}{c|}{Superpoint} & \multicolumn{2}{c|}{R2D2} & \multicolumn{2}{c|}{DISK} & \multicolumn{2}{c}{Ours} \\
          & Kpts  & MMA5  & Kpts  & MMA5  & Kpts  & MMA5  & Kpts  & MMA5  & Kpts  & MMA5 \\
    \midrule
    002\_master\_chef\_can & 20.5  & 35.8\% & 25.8  & \textbf{48.8\%} & 30.4  & 46.5\% & 15.9  & 46.6\% & \textbf{187.0 } & 41.7\% \\
    003\_cracker\_box & 44.6  & 73.9\% & 48.6  & 81.2\% & 70.1  & 81.0\% & 68.1  & \textbf{86.2\%} & \textbf{176.6 } & 74.7\% \\
    004\_sugar\_box & 38.4  & 63.4\% & 42.4  & 69.2\% & 54.5  & 65.9\% & 50.4  & 73.0\% & \textbf{229.0 } & \textbf{73.6\%} \\
    005\_tomato\_soup\_can & 32.5  & 62.7\% & 24.6  & 68.7\% & 26.5  & 71.1\% & 19.5  & \textbf{71.9\%} & \textbf{177.8 } & 70.4\% \\
    006\_mustard\_bottle & 12.8  & 24.2\% & 19.5  & 48.2\% & 23.3  & 41.6\% & 17.3  & 38.6\% & \textbf{123.2 } & \textbf{50.3\%} \\
    007\_tuna\_fish\_can & 10.4  & 37.1\% & 7.6   & \textbf{61.6\%} & 5.8   & 40.3\% & 3.3   & 2.9\% & \textbf{23.1 } & 29.1\% \\
    008\_pudding\_box & 30.2  & 1.8\% & 18.0  & 15.3\% & 21.6  & 70.8\% & 16.3  & 18.3\% & \textbf{77.5 } & \textbf{87.5\%} \\
    009\_gelatin\_box & 74.6  & 86.1\% & 42.6  & 83.0\% & 57.8  & 77.8\% & 41.5  & \textbf{89.3\%} & \textbf{259.5 } & 82.8\% \\
    010\_potted\_meat\_can & 19.0  & 36.5\% & 14.8  & 39.4\% & 19.8  & 47.3\% & 17.4  & 49.7\% & \textbf{122.5 } & \textbf{61.5\%} \\
    011\_banana & 3.7   & 0.3\% & 2.4   & 0.2\% & 4.0   & 0.4\% & 2.9   & 0.0\% & \textbf{15.4 } & \textbf{14.7\%} \\
    019\_pitcher\_base & 4.5   & 0.3\% & 4.3   & 11.8\% & 4.7   & 5.3\% & 4.0   & 2.6\% & \textbf{18.4 } & \textbf{35.0\%} \\
    021\_bleach\_cleanser & 17.5  & 24.4\% & 22.8  & 26.6\% & 37.0  & 25.9\% & 30.1  & 24.5\% & \textbf{69.8 } & \textbf{34.1\%} \\
    024\_bowl & 0.9   & 0.0\% & 2.2   & 0.0\% & 5.1   & 0.5\% & 4.9   & 0.5\% & \textbf{16.2 } & \textbf{6.9\%} \\
    025\_mug & 4.2   & 0.0\% & 3.3   & 0.9\% & 4.8   & 1.3\% & 4.7   & 1.4\% & \textbf{60.8 } & \textbf{29.3\%} \\
    035\_power\_drill & 6.8   & 8.0\% & 20.8  & 69.2\% & 26.1  & \textbf{71.6\%} & 13.2  & 67.2\% & \textbf{64.3 } & 58.0\% \\
    036\_wood\_block & 5.3   & 0.7\% & 2.9   & 0.0\% & 3.7   & 0.0\% & 4.1   & 0.0\% & \textbf{13.7 } & \textbf{4.7\%} \\
    037\_scissors & 4.3   & 4.4\% & 3.3   & 2.0\% & 2.9   & 2.3\% & 3.0   & 0.7\% & \textbf{44.6 } & \textbf{85.6\%} \\
    040\_large\_marker & 11.8  & 48.9\% & 11.7  & 66.9\% & 7.4   & 55.6\% & 3.9   & 16.5\% & \textbf{85.2 } & \textbf{70.4\%} \\
    051\_large\_clamp & 5.6   & 0.2\% & 3.5   & 2.0\% & 4.5   & 5.5\% & 3.7   & 0.1\% & \textbf{32.0 } & \textbf{39.5\%} \\
    052\_extra\_large\_clamp & 6.1   & 0.2\% & 4.6   & 5.3\% & 4.8   & 6.5\% & 4.3   & 0.4\% & \textbf{36.2 } & \textbf{19.4\%} \\
    061\_foam\_brick & 1.2   & 0.0\% & 3.2   & 4.7\% & 4.0   & 12.8\% & 3.1   & 0.1\% & \textbf{101.5 } & \textbf{78.6\%} \\
    \midrule
    Average & 16.9  & 24.2\% & 15.7  & 33.6\% & 20.0  & 34.8\% & 15.8  & 28.1\% & \textbf{92.1}  & \textbf{50.0\%} \\
    \end{tabular}
    
}%
  \label{tab:syn-real}%
\end{table*}%

\subsubsection{Visualization of real-real image matching}
Due to the apparent movement of the target object in the scene, the matching performance of scene-centric methods, which ignore the objectness of the keypoints, deteriorates significantly. In contrast, our method can still detect some matching points when the current frame has a large deviation from the initial frame, as shown in Figure~\ref{fig:real_real2}. More intuitive performance can be seen in the supplementary video.

\subsubsection{6D pose estimation}
\begin{figure*}[htbp]
    \centering
    \includegraphics[width=0.99\textwidth]{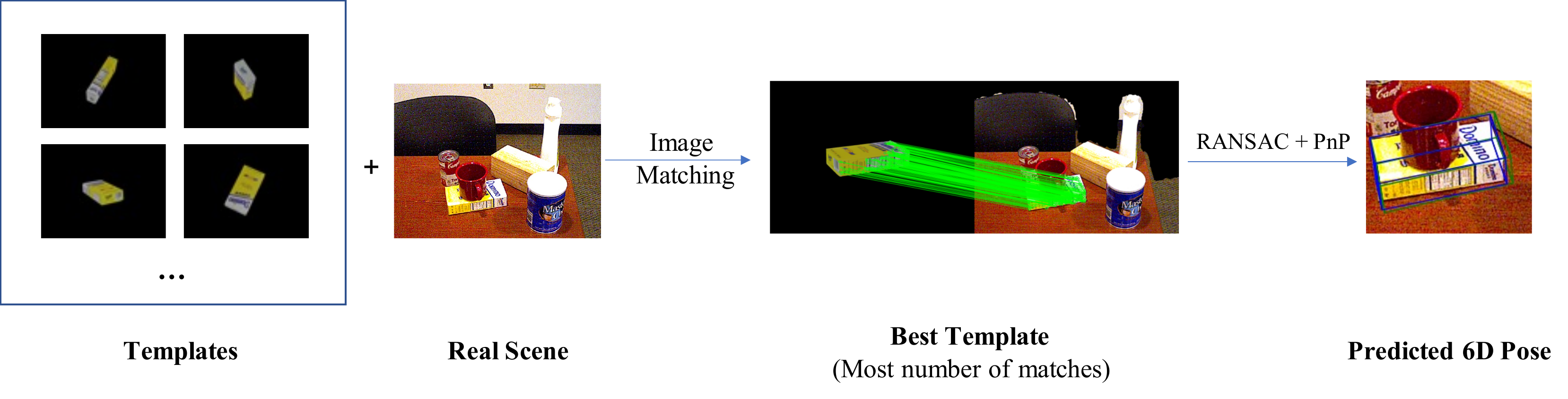}
    \caption{Pose estimation pipeline.}
    \label{fig:pose_pipeline}
\end{figure*}

The flow chart of unsupervised pose estimation is shown in Figure~\ref{fig:pose_pipeline}: (1) Given a real image and a query object $i$, we first render multiple images in different poses of object $i$ as templates. (2) Match the templates with the real image one by one, and select the best template according to the number of matched pairs. (3) The object pose can be solved through the Perspective-n-Point (PnP) and RANSAC algorithms. However, if the number of matched points is below 4, we conclude that the object $i$ does not appear in this scene.


\begin{figure*}[!ht]
    \centering
    \includegraphics[width=0.9\textwidth]{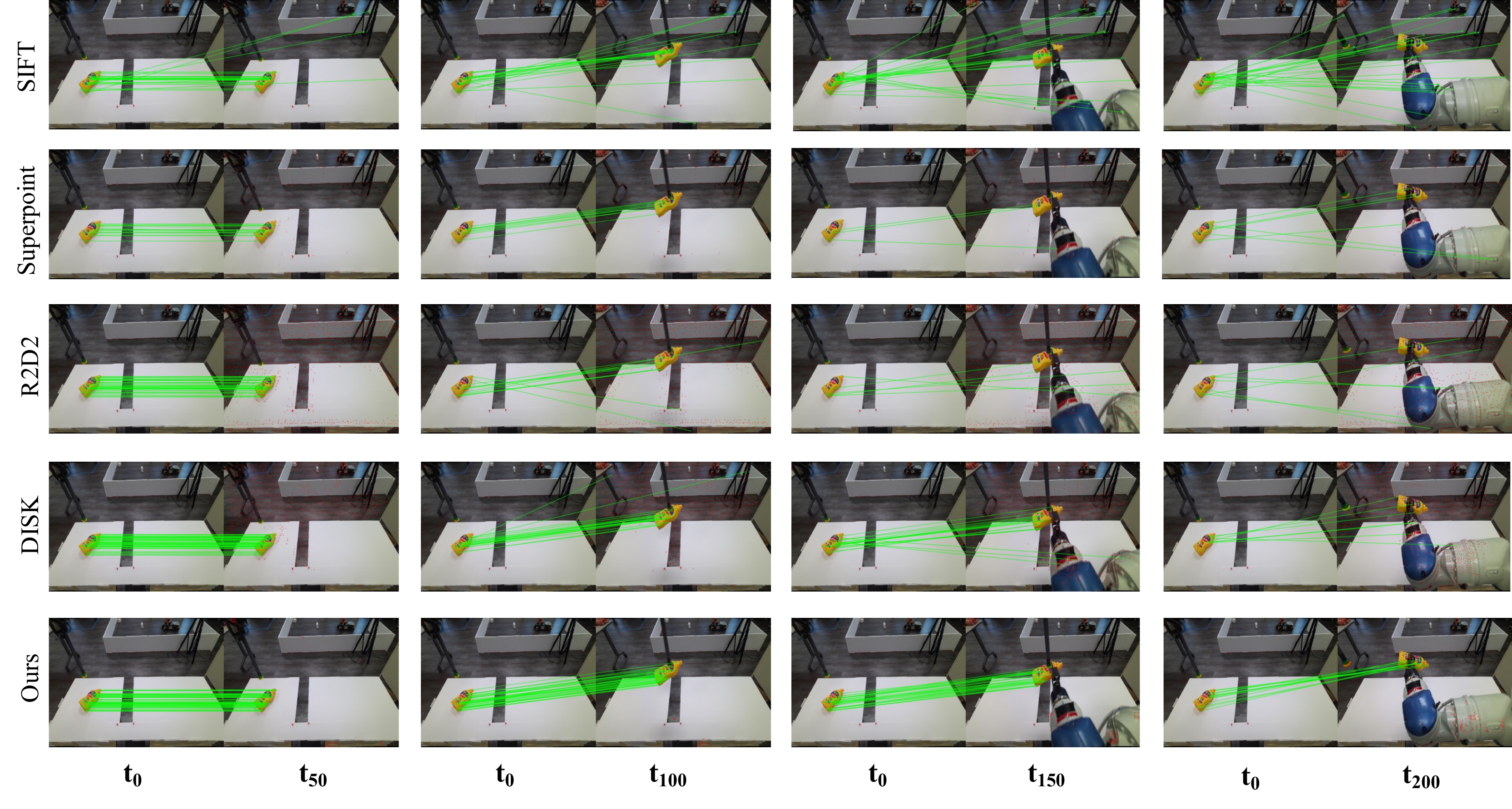}
    \vskip -0.1in
    \caption{Qualitative results of real-real image matching over time. $t_k$=$k$-th frame. The performance of other objects can be seen in the supplementary videos.}
    \label{fig:real_real2}
\end{figure*}

\subsection{Additional Ablation Experiments}
\noindent
\subsubsection{Sensitivity to temperature parameter $\tau$} 

We design four groups of parameters comparative experiments on temperature  $\tau$ for learning two subparts of intra-object salience and inter-object distinctness descriptors (parameters of InfoNCE~\cite{infonce} loss), as illustrated in Table~\ref{tab:more_dim}. We choose the parameters, i.e., intra-object $\tau=0.07$ and inter-object $\tau=0.2$ of the InfoNCE for the final performance.

\begin{table}[!ht]
  \centering
  \caption{Sensitivity to temperature $\tau$. obj=object, syn=synthetic.}
  \vspace{8pt}
  \resizebox{0.98\columnwidth}{!}{
    \begin{tabular}{cc|cccc}
    \multicolumn{1}{c}{Intra-obj} & \multicolumn{1}{c|}{Inter-obj} & Syn-Real & Real-Real & \multicolumn{2}{c}{6D Pose} \\
    \multicolumn{1}{c}{$\tau$} & \multicolumn{1}{c|}{$\tau$} & MMA5  & MMA5  & ADD   & ADDS \\
    \midrule
    0.07  & 0.07  & 43.6\% & 38.6\% & 40.2  & 59.5  \\
    0.07  & 0.2   & \textbf{50.0\%} & \textbf{38.8\%} & \textbf{42.2} & \textbf{61.9} \\
    0.2   & 0.07  & 37.7\% & 34.8\% & 32.1  & 50.1  \\
    0.2   & 0.2   & 35.6\% & 36.3\% & 30.2  & 48.5  \\
    \end{tabular}}%
  \label{tab:more_dim}%
\end{table}%

\noindent
\subsubsection{Various dimensionality of keypoint descriptor}
In Table~\ref{tab:more_descriptor}, we evaluate the impact of two subparts of keypoint descriptor on various dimensions. It is suggested that we need to use more parameters to distinguish differences in intra-objects. Due to the small number of objects in the training set, too many inter-object representation vectors may lead to model overfitting.

\begin{table}[!ht]
  \centering
  \caption{Ablation study on the dimension of descriptors. obj=object, syn=synthetic, dim=dimensions.}
  \vspace{8pt}
    \resizebox{0.95\columnwidth}{!}{
    \begin{tabular}{cc|cccc}
    \multicolumn{1}{c}{Intra-obj} & \multicolumn{1}{c|}{Inter-obj} & Syn-Real & Real-Real & \multicolumn{2}{c}{6D Pose} \\ 
    \multicolumn{1}{c}{Dim} & \multicolumn{1}{c|}{Dim} & \small{MMA5}  & \small{MMA5}  & \small{ADD}   & \small{ADDS}  \\ 
    \midrule
    32    & 32    & 43.3\% & 32.6\% & 36.1  & 54.0  \\ 
    64    & 32    & \textbf{50.0\%} & \textbf{38.8\%} & \textbf{41.1} & \textbf{61.9} \\ 
    32    & 64    & 44.0\% & 37.2\% & 38.1  & 58.2  \\ 
    64    & 64    & 44.7\% & 36.2\% & 39.0  & 58.2  \\ 
    \end{tabular}}
  \label{tab:more_descriptor}%
\end{table}%

\noindent
\subsubsection{The number of templates}
Because the number of templates is essential for 6D pose estimation, we have conducted the ablation study by setting 48, 72, and 96 templates, whose ADDS are 55.8, 57.7, and 61.9, respectively, which implies the optimal choice of our setting (\emph{i.e.} 96). More templates will bring better pose estimation results, but correspondingly will reduce the computation efficiency, so we recommend that the number of templates is fewer than 100.

\noindent
\subsubsection{Matching strategy}
Because of the explicit objectness descriptors, there are two template matching strategies. The first is a non-direct matching method. We exploit the mean inter-object descriptor of the template as the key and find the pixels in a real image similar to the key. Then, the matching is performed according to the intra-object descriptors of the template and the selected pixels in the real image. The $cosine$ measure is used when calculating the inter-object similarity, and the similarity threshold is 0.5. 
In contrast, the direct method is that the intra-object and inter-object descriptors are directly concatenated as the general descriptor of each pixel. The correspondence between the template and a real image can be found according to the general descriptors. The results of the two strategies are shown in Table~\ref{tab:match_strategy}. Obviously, the explicit objectness can help filter the similar geometric or semantic points in different objects between the template and the real image to get better matching effects.

\begin{table}[!ht]
  \centering
  \caption{The ablation study of the two matching strategies.}
    \resizebox{0.85\columnwidth}{!}{
    \begin{tabular}{c|c|c|cc}
    \multirow{2}[1]{*}{Direct} & Syn-Real & Real-Real & \multicolumn{2}{c}{6D Pose} \\
          & MMA5  & MMA5  & ADD   & ADDS \\
    \midrule
    \checkmark   & 53.6\% & 31.5\% & 30.5  & 46.3  \\
          & 50.0\% & 38.8\% & 42.2  & 61.9  \\
    \end{tabular}}%
  \label{tab:match_strategy}%
\end{table}%

\subsection{Model Generalization}
\begin{figure*}[htbp]
    \centering
    \includegraphics[width=0.7\textwidth]{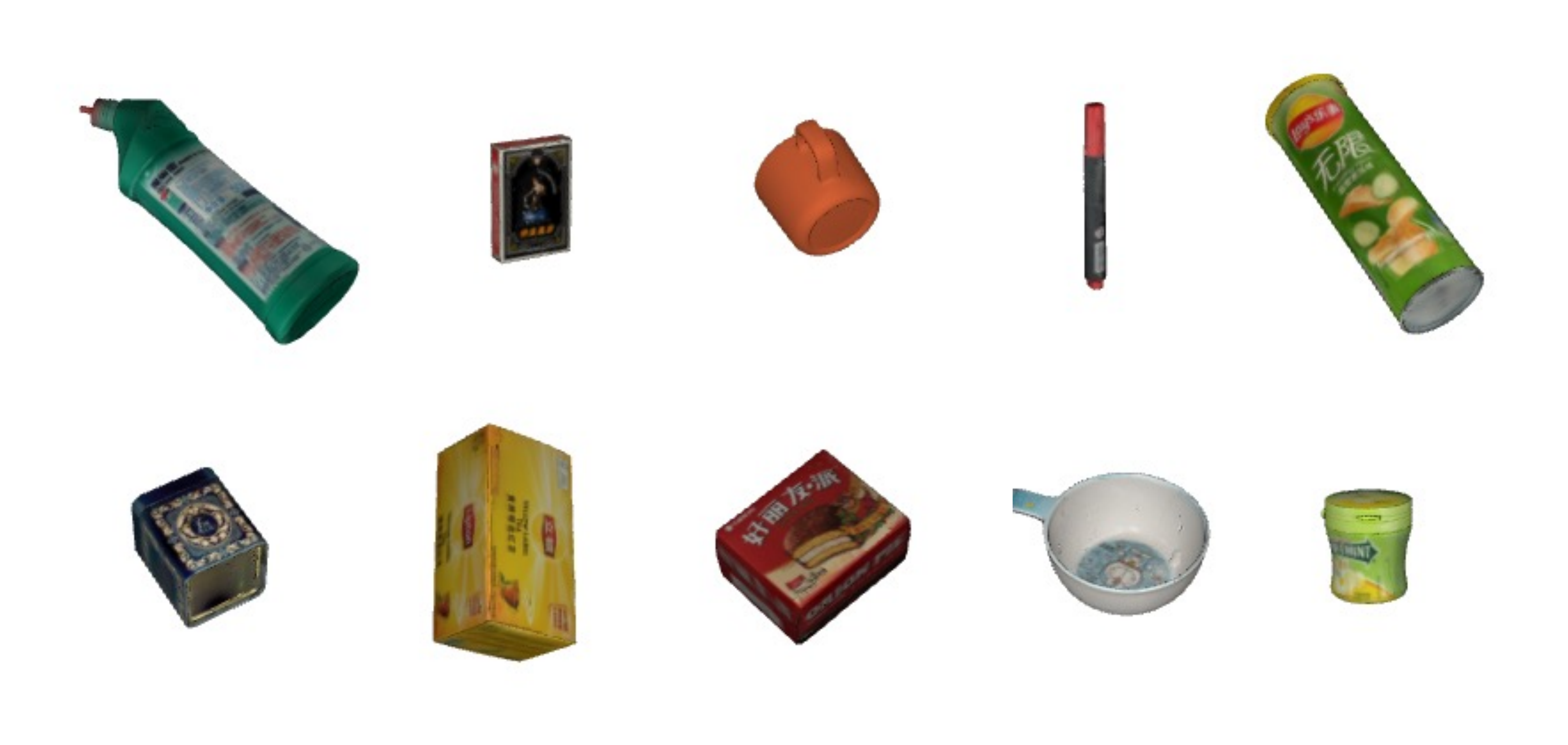}
    \caption{The ten novel objects with the same class as YCB-Video objects, like bowl, bottle and box.}
    \label{fig:seen_class}
\end{figure*}

\begin{figure*}[htbp]
    \centering
    \includegraphics[width=0.7\textwidth]{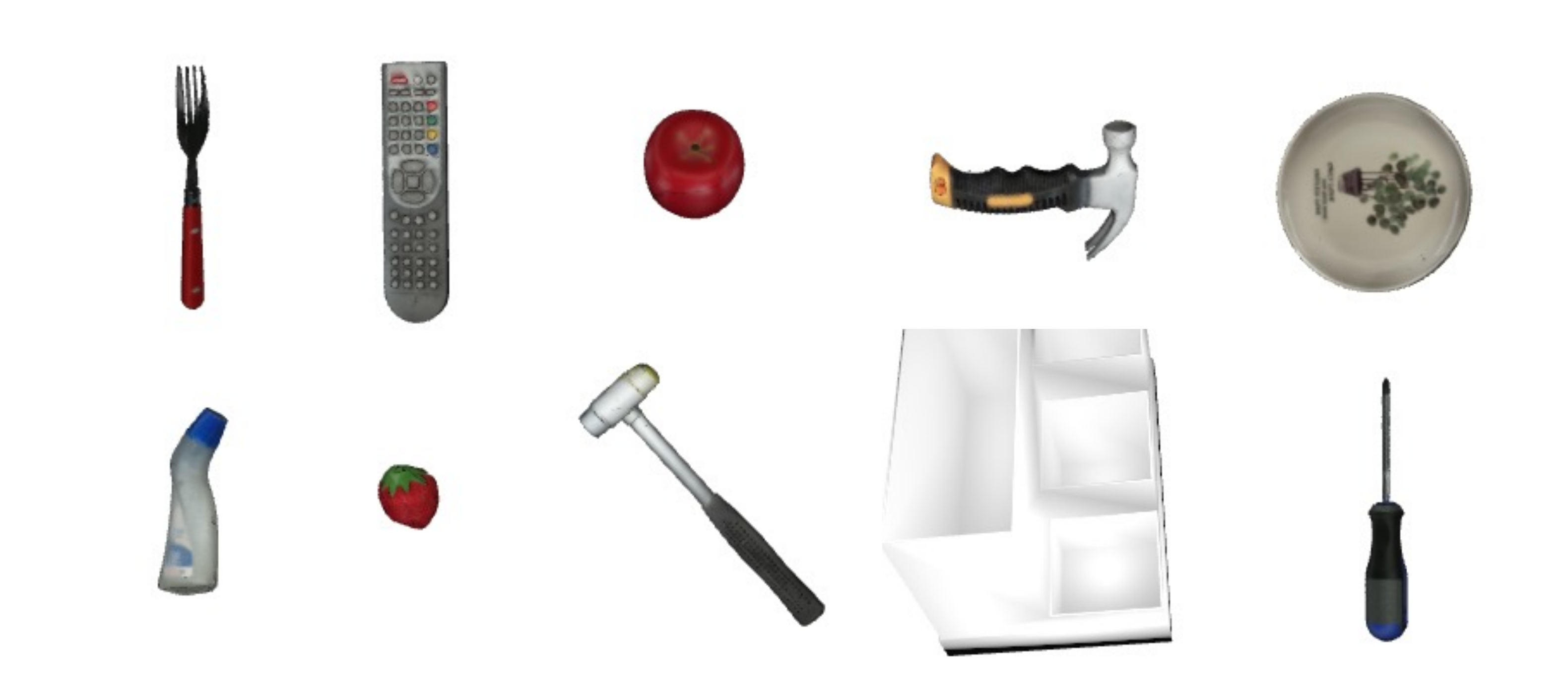}
    \caption{The ten novel objects with different class from YCB-Video objects, like hardware tool, book holder and silverware.}
    \label{fig:unseen_class}
\end{figure*}

Twenty unseen objects are selected from the OCRTOC dataset~\cite{liu2021ocrtoc}, in which half of them have the same class as YCB-Video objects but with different shapes or textures (seen class), which can be seen in Figure~\ref{fig:seen_class}. The left with novel classes has not been seen in training (unseen class), which are visualized in Figure~\ref{fig:unseen_class}. We generate the simulated images for unseen objects under new backgrounds (textures) which have not been seen in training. Objects are translated and rotated slightly while the rendered camera is static. The evaluation protocol is similar to real-real matching. 

\begin{figure*}[htbp]
    \centering
    \includegraphics[width=0.9\textwidth]{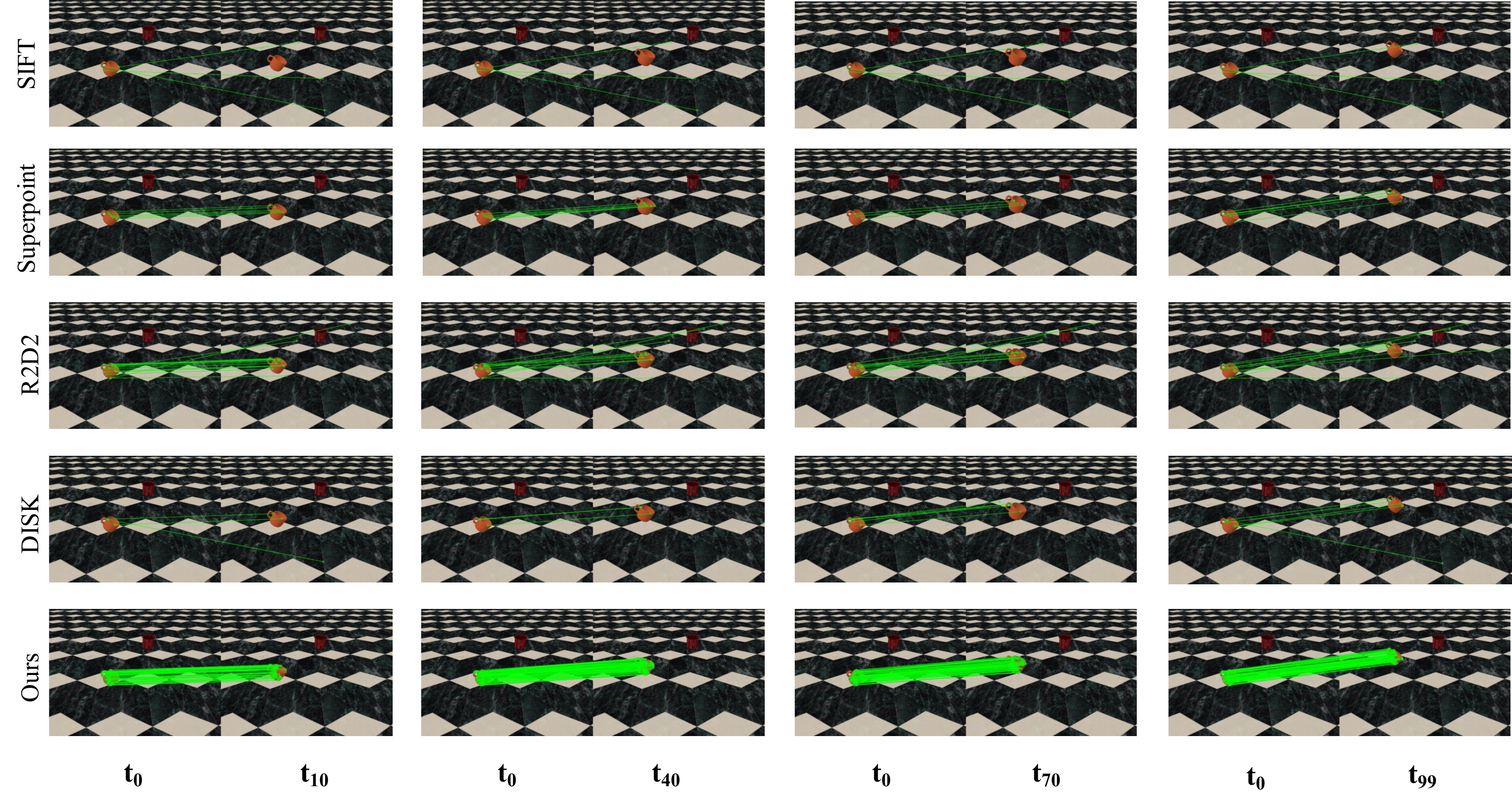}
    \caption{Qualitative results of image matching of the object \textit{orange box} over time. $t_k$=$k$-th frame.}
    \label{fig:unseen_orange}
\end{figure*}

\begin{figure*}[htbp]
    \centering
    \includegraphics[width=0.95\textwidth]{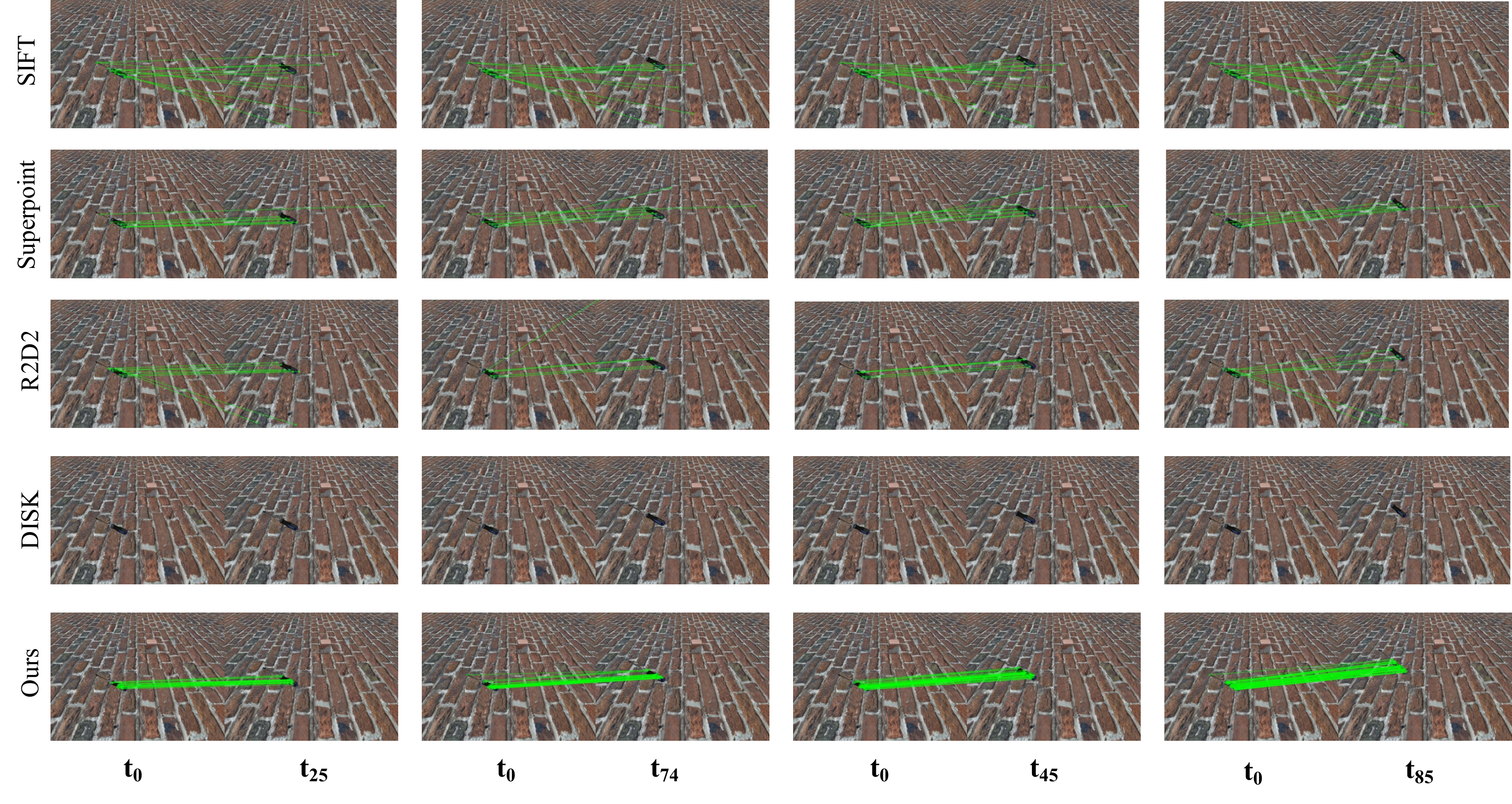}
    \caption{Qualitative results of image matching of the object \textit{phillips screwdrive} over time. $t_k$=$k$-th frame. }
    \label{fig:unseen_philips}
\end{figure*}

From Table~\ref{tab:unseen_match1} and \ref{tab:unseen_match2}, we can see that our method also achieves satisfactory matching performance for unseen objects. Because we can first find the objectness of each keypoint to restrict the set of potential matching points, then further obtain corresponding points by comparing their intra-descriptors. While other descriptors ignore the objectness, there will be more mismatch points. The visualization of unseen objects matching results is shown in Figure~\ref{fig:unseen_orange} and ~\ref{fig:unseen_philips}.

\begin{table*}[!ht]
  \centering
  \caption{Evaluation for image matching on unseen objects (seen class).}
    \resizebox{0.9\textwidth}{!}{
    \begin{tabular}{cc|cc|cc|cc|cc|cc}
    \multicolumn{1}{l}{\multirow{2}[1]{*}{Objects}} & \multicolumn{1}{l|}{\multirow{2}[1]{*}{Class}} & \multicolumn{2}{c|}{SIFT} & \multicolumn{2}{c|}{Superpoint} & \multicolumn{2}{c|}{R2D2} & \multicolumn{2}{c|}{DISK} & \multicolumn{2}{c}{Ours} \\
          &       & Kpts  & MMA5  & Kpts  & MMA5  & Kpts  & MMA5  & Kpts  & MMA5  & Kpts  & MMA5 \\
    \midrule
    \multicolumn{1}{l}{magic\_clean} & \multicolumn{1}{l|}{bottle} & \textbf{46.6 } & 36.5\% & 19.4  & 61.5\% & 30.1  & 74.9\% & 40.0  & \textbf{89.2\%} & 30.8  & 66.6\% \\
    \multicolumn{1}{l}{poker} & \multicolumn{1}{l|}{box} & 16.4  & 37.0\% & 6.2   & \textbf{97.9\%} & 15.2  & 59.0\% & 15.6  & 81.2\% & \textbf{60.7 } & 93.5\% \\
    \multicolumn{1}{l}{orange\_cup} & \multicolumn{1}{l|}{cup} & 3.0   & 0.0\% & 4.5   & 23.1\% & 10.8  & 18.7\% & 4.4   & 22.3\% & \textbf{100.0 } & \textbf{40.4\%} \\
    \multicolumn{1}{l}{red\_marker} & \multicolumn{1}{l|}{marker} & \textbf{14.8 } & 2.1\% & 1.2   & 0.0\% & 6.4   & 3.6\% & 2.0   & 3.8\% & \textbf{14.8 } & \textbf{75.8\%} \\
    \multicolumn{1}{l}{potato\_chip\_3} & \multicolumn{1}{l|}{can} & 17.8  & 33.6\% & 20.9  & 60.5\% & 26.5  & \textbf{86.5\%} & 36.6  & 60.5\% & \textbf{101.3 } & 72.1\% \\
    \multicolumn{1}{l}{blue\_tea\_box} & \multicolumn{1}{l|}{can} & 18.1  & 31.9\% & 4.5   & 28.4\% & 16.4  & 75.2\% & 15.2  & 44.6\% & \textbf{81.9 } & \textbf{90.7\%} \\
    \multicolumn{1}{l}{lipton\_tea} & \multicolumn{1}{l|}{box} & 28.8  & 72.5\% & 36.7  & 90.4\% & 43.4  & 82.9\% & 41.6  & \textbf{90.5\%} & \textbf{220.1 } & 85.8\% \\
    \multicolumn{1}{l}{orion\_pie} & \multicolumn{1}{l|}{box} & 53.8  & 65.1\% & 28.7  & 87.3\% & 29.1  & 86.7\% & 21.7  & 78.7\% & \textbf{196.1 } & \textbf{91.1\%} \\
    \multicolumn{1}{l}{doraemon\_bowl} & \multicolumn{1}{l|}{bowl} & 14.7  & 14.5\% & 5.7   & 40.1\% & 22.4  & \textbf{44.3\%} & 9.7   & 26.0\% & \textbf{94.3 } & 22.0\% \\
    \multicolumn{1}{l}{sugar\_1} & \multicolumn{1}{l|}{bottole} & \textbf{19.3 } & 32.1\% & 11.4  & 61.4\% & 11.7  & 80.8\% & 17.1  & \textbf{82.6\%} & 5.0   & 16.3\% \\
    \midrule
    \multicolumn{2}{c|}{Average} & 23.3  & 32.5\% & 13.9  & 55.0\% & 21.2  & 61.3\% & 20.4  & 57.9\% & \textbf{90.5 } & \textbf{65.4\%} \\
\end{tabular}
 }%
  \label{tab:unseen_match1}%
\end{table*}%

\begin{table*}[!ht]
  \centering
  \caption{Evaluation for image matching on unseen objects (unseen class).}
    \resizebox{0.95\textwidth}{!}{
    \begin{tabular}{cc|cc|cc|cc|cc|cc}
    \multicolumn{1}{l}{\multirow{2}[1]{*}{Objects}} & \multicolumn{1}{l|}{\multirow{2}[1]{*}{Class}} & \multicolumn{2}{c|}{SIFT} & \multicolumn{2}{c|}{Superpoint} & \multicolumn{2}{c|}{R2D2} & \multicolumn{2}{c|}{DISK} & \multicolumn{2}{c}{Ours} \\
          &       & Kpts  & MMA5  & Kpts  & MMA5  & Kpts  & MMA5  & Kpts  & MMA5  & Kpts  & MMA5 \\
    \midrule
    \multicolumn{1}{l}{fork} & \multicolumn{1}{l|}{silverware} & 8.2   & 30.6\% & 4.4   & 16.6\% & 4.2   & 6.9\% & 4.3   & 27.1\% & \textbf{36.1 } & \textbf{46.0\%} \\
    \multicolumn{1}{l}{remote\_controller\_2} & \multicolumn{1}{l|}{remote} & 36.7  & 16.3\% & 9.6   & 48.7\% & 13.9  & 35.2\% & 8.1   & 30.3\% & \textbf{55.3 } & \textbf{52.8\%} \\
    \multicolumn{1}{l}{plastic\_apple} & \multicolumn{1}{l|}{fruit} & 8.2   & 34.9\% & 6.4   & 41.5\% & 6.2   & \textbf{80.4\%} & 6.9   & 49.7\% & \textbf{78.7 } & 59.8\% \\
    \multicolumn{1}{l}{mini\_claw\_hammer\_1} & \multicolumn{1}{l|}{hardware\_tools} & 33.7  & 21.7\% & 14.5  & 71.5\% & 15.5  & 73.8\% & 7.1   & 51.3\% & \textbf{45.6 } & \textbf{82.5\%} \\
    \multicolumn{1}{l}{round\_plate\_1} & \multicolumn{1}{l|}{plate} & 18.0  & 43.9\% & 16.2  & 60.6\% & 18.9  & \textbf{72.1\%} & 8.9   & 57.3\% & \textbf{105.6 } & 47.5\% \\
    \multicolumn{1}{l}{glue\_2} & \multicolumn{1}{l|}{stationery} & 8.0   & 25.3\% & 8.7   & 70.9\% & 20.4  & \textbf{89.5\%} & 5.9   & 78.5\% & \textbf{88.3 } & 60.0\% \\
    \multicolumn{1}{l}{plastic\_strawberry} & \multicolumn{1}{l|}{fruit} & 13.9  & 38.8\% & 1.0   & 0.0\% & 3.2   & 2.1\% & 4.2   & 36.4\% & \textbf{18.5 } & \textbf{73.8\%} \\
    \multicolumn{1}{l}{two\_color\_hammer} & \multicolumn{1}{l|}{hardware\_tools} & 16.3  & 47.1\% & 7.0   & \textbf{71.7\%} & 12.1  & 60.9\% & 3.5   & 29.3\% & \textbf{68.7 } & 69.9\% \\
    \multicolumn{1}{l}{book\_holder} & \multicolumn{1}{l|}{stationery} & 13.8  & 10.6\% & 8.5   & \textbf{41.4\%} & 60.4  & 26.5\% & 6.9   & 3.1\% & \textbf{226.6 } & 29.3\% \\
    \multicolumn{1}{l}{phillips\_screwdriver} & \multicolumn{1}{l|}{\textcolor[rgb]{ .141,  .161,  .184}{hardware\_tools}} & 11.2  & 26.5\% & 7.3   & 60.8\% & 6.4   & 45.8\% & 0.4   & 1.0\% & \textbf{35.3 } & \textbf{91.7\%} \\
    \midrule
    \multicolumn{2}{c|}{Average} & 16.8  & 29.6\% & 8.3   & 48.30\% & 16.1  & 49.3\% & 5.6   & 36.4\% & \textbf{75.9 } & \textbf{61.3\%} \\
\end{tabular}%

 }%
  \label{tab:unseen_match2}%
\end{table*}%

	\end{appendix}
	
\end{document}